\documentclass[11pt]{article}
\usepackage{makecell}
\usepackage[preprint]{acl}
\usepackage{booktabs}
\usepackage{amsmath}
\usepackage{mathtools}
\usepackage{multirow}
\usepackage{caption} 
\usepackage{times}
\usepackage{latexsym}
\usepackage[table]{xcolor}
\usepackage[breakable,skins]{tcolorbox}

\newcommand*\justify{%
  \fontdimen2\font=0.4em
  \fontdimen3\font=0.2em
  \fontdimen4\font=0.1em
  \fontdimen7\font=0.1em
  \hyphenchar\font=`\-
}

\renewcommand{\texttt}[1]{%
\begingroup
\ttfamily
\begingroup\lccode`~=`/\lowercase{\endgroup\def~}{/\discretionary{}{}{}}%
\begingroup\lccode`~=`[\lowercase{\endgroup\def~}{[\discretionary{}{}{}}%
\begingroup\lccode`~=`.\lowercase{\endgroup\def~}{.\discretionary{}{}{}}%
\catcode`/=\active\catcode`[=\active\catcode`.=\active
\justify\scantokens{#1\noexpand}%
\endgroup
}

\PassOptionsToPackage{numbers, compress}{natbib}

\usepackage[utf8]{inputenc}         %
\usepackage[T1]{fontenc}            %
\usepackage{graphicx}
\usepackage{todonotes}
\usepackage{multirow}
\usepackage{hyperref}
\usepackage[capitalize,noabbrev,nameinlink,sort]{cleveref}
\usepackage{subcaption}
\usepackage{multicol}
\usepackage{array}
\usepackage{enumitem}
\usepackage{algorithm}
\usepackage{algpseudocode}
\usepackage{tabularx}
\usepackage{listings}
\usepackage{makecell}
\usepackage{xspace}
\usepackage{colortbl}
\usepackage{fontawesome5}
\usepackage{pifont}
\usepackage[normalem]{ulem}
\useunder{\uline}{\ul}{}
\usepackage{tablefootnote}
\usepackage{tcolorbox}
\usepackage{arydshln}
\usepackage[toc, page, header]{appendix}
\usepackage{tabularx}
\usepackage{listings}

\usepackage{amsmath}
\usepackage{amssymb}
\usepackage{amsfonts}                               
\usepackage{amsthm}
\usepackage[mathcal]{eucal}
\usepackage{mathrsfs}
\usepackage{bm}                                     
\usepackage{blkarray}                               
\usepackage{nicefrac}                               
\usepackage{bbm}

\usepackage{wrapfig}
\usepackage{graphicx}                               
\usepackage{caption}
\usepackage{tikz}
\usepackage{circuitikz}
\usetikzlibrary{patterns,snakes}
\usetikzlibrary{positioning,calc,fit,decorations.pathmorphing,shapes.geometric, shapes.gates.logic.US, calc}
\usetikzlibrary{arrows,arrows.meta,decorations.markings,shapes,shapes.arrows}
\usetikzlibrary{decorations,decorations.pathreplacing}
\usetikzlibrary{backgrounds}
\usepackage{filecontents}                           
\usepackage{pgfplots}
\usepackage{pgfplotstable}
\usepgfplotslibrary{groupplots}
\usepackage{scalefnt}
\pgfplotsset{compat=newest}
\usepackage{xcolor}

\definecolor{firstcolor}{HTML}{C3423F}
\definecolor{secondcolor}{HTML}{2A4B8C}
\definecolor{aworld_blue}{HTML}{4e81ff}
\definecolor{aworld_cyan}{HTML}{41d7fa}
\definecolor{aworld_teal}{HTML}{5fede4}
\definecolor{coral}{RGB}{255,127,80}
\definecolor{darkgreen}{RGB}{0,100,0}
\definecolor{darkyellow}{RGB}{204,153,0}
\definecolor{salmon}{RGB}{250,128,114}
\definecolor{darkred}{RGB}{150,0,0}

\hypersetup{
  colorlinks=true,
  linkbordercolor=aworld_blue,
  citebordercolor=aworld_cyan,
  urlbordercolor=aworld_teal
}

\definecolor{improvementblue}{RGB}{55,126,184}    
\definecolor{degradationorange}{RGB}{230,85,13}   

\def\eqref#1{equation~\ref{#1}}

\def\1{\bm{1}}

\DeclareMathAlphabet{\mathsfit}{\encodingdefault}{\sfdefault}{m}{sl}
\SetMathAlphabet{\mathsfit}{bold}{\encodingdefault}{\sfdefault}{bx}{n}

\usepackage[T1]{fontenc}
\usepackage{xcolor}
\usepackage{pifont}
\usepackage{booktabs}
\usepackage{amsmath}
\usepackage{bbm}

\usepackage[utf8]{inputenc}
\usepackage{subcaption}
\usepackage{microtype}
\usepackage{bbding}
\usepackage{inconsolata}
\usepackage{graphicx}

\title{}
\author{}

\begin{document}

\twocolumn[{%
\noindent\includegraphics[height=1.0cm]{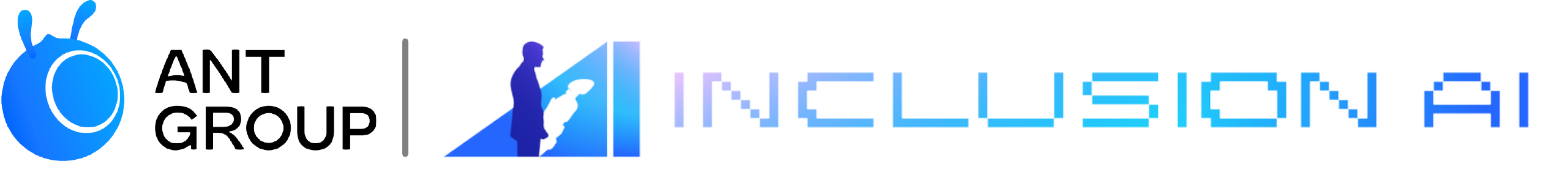}\\[0.3em]
\noindent\rule{\textwidth}{0.4pt}

\vspace{0.5em}
\begin{center}
{\Large\textbf{From Failure to Mastery: Generating Hard Samples for Tool-use Agents}}\\[0.2em]
\end{center}
\vspace{0.4em}

\noindent\rule{\textwidth}{0.4pt}

\vspace{0.6em}
\begin{center}
\textbf{Bingguang Hao\textsuperscript{1*}}, \textbf{Zengzhuang Xu\textsuperscript{1*}}, \textbf{Yuntao Wen\textsuperscript{1*}}, \textbf{Xinyi Xu\textsuperscript{1*}}, \textbf{Yang Liu\textsuperscript{2*}},
\\[0.3em]
\textbf{Tong Zhao\textsuperscript{2}}, \textbf{Maolin Wang\textsuperscript{3}}, \textbf{Long Chen\textsuperscript{1}}, \textbf{Dong Wang\textsuperscript{1}}, \textbf{Yicheng Chen\textsuperscript{1}},
\\[0.3em]
\textbf{Cunyin Peng\textsuperscript{1}}, \textbf{Xiangyu Zhao\textsuperscript{3}}, \textbf{Chenyi Zhuang\textsuperscript{1‡}}, \textbf{Ji Zhang\textsuperscript{4‡}}
\\[0.5em]
\small
\textsuperscript{1}Inclusion AI, Ant Group \quad
\textsuperscript{2}Zhejiang University
\\
\textsuperscript{3}City University of Hong Kong \quad
\textsuperscript{4}Southwest Jiaotong University
\\[0.3em]
\texttt{\{bingguanghao7,jizhang.jim\}@gmail.com} \quad
\texttt{\{chenyi.zcy\}@antgroup.com}
\\[0.3em]
\raisebox{-0.1em}{\includegraphics[height=0.9em]{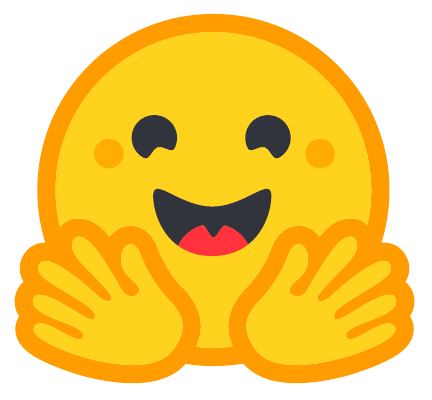}} \href{https://huggingface.co/datasets/Bingguang/FunReason-MT}{\texttt{Dataset}}\quad\quad\faGithub\ \href{https://github.com/inclusionAI/AWorld-RL}{\texttt{HardGen}}
\end{center}

\vspace{0.8em}
}]

{\renewcommand\thefootnote{}
\footnotetext{*Equal contributions. ‡Corresponding Authors.}}

\begin{abstract}
The advancement of LLM agents with tool-use capabilities requires diverse and complex training corpora. 
Existing data generation methods, which predominantly follow a paradigm of random sampling and shallow generation, often yield simple and homogeneous trajectories that fail to capture complex, implicit logical dependencies.
To bridge this gap, we introduce \textbf{HardGen}, an automatic agentic pipeline designed to generate hard tool-use training samples with verifiable reasoning.  
\textit{Firstly}, HardGen establishes a dynamic API Graph built upon agent failure cases, from which it samples to synthesize hard traces.
\textit{Secondly}, these traces serve as conditional priors to guide the instantiation of modular, abstract advanced tools, which are subsequently leveraged to formulate hard queries.
\textit{Finally}, the advanced tools and hard queries enable the generation of verifiable complex Chain-of-Thought (CoT), with a closed-loop evaluation feedback steering the continuous refinement of the process.
Extensive evaluations demonstrate that a 4B parameter model trained with our curated dataset achieves superior performance compared to several leading open-source and closed-source competitors (\textit{e.g.}, GPT-5.2, Gemini-3-Pro and Claude-Opus-4.5).
Our code, models, and dataset will be open-sourced to facilitate future research.

\end{abstract}


\section{Introduction}
\label{sec:intro}


\begin{figure}[t]
    \centering
    \includegraphics[width=0.9\linewidth]{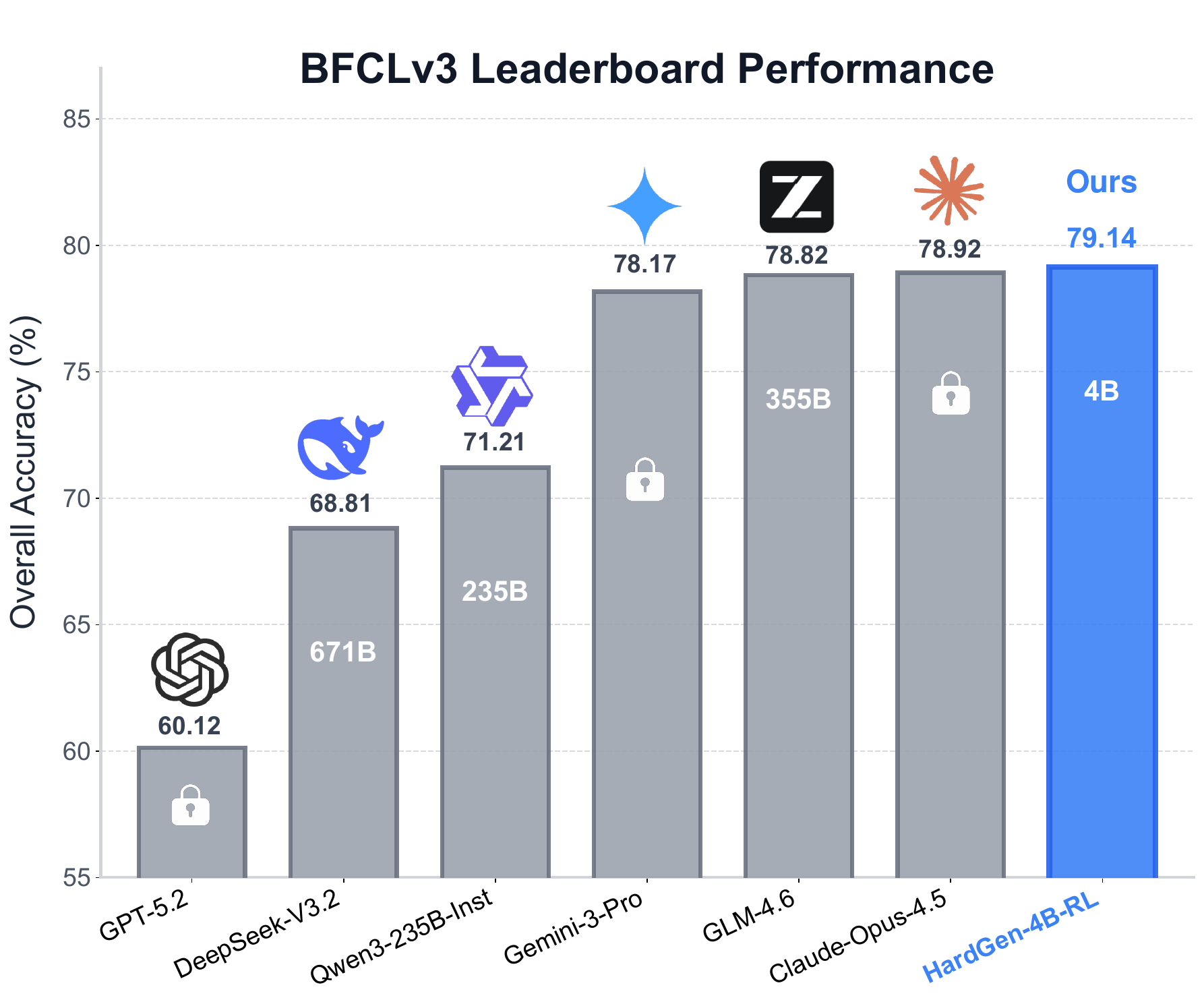}
    \caption{\textbf{Performance comparison on the BFCLv3} \textbf{Leaderboard} \cite{patilberkeley}. A Qwen3-4B~\cite{yang2025qwen3} model trained with our curated dataset, denoted as \textbf{HardGen-4B-RL},
    consistently outperforms leading open-source and closed-source models.
    }
    \label{fig:perfo1}
\end{figure}


Equipping Large Language Models (LLMs) with the capability to execute external tools\footnote{We use tools and APIs interchangeably in this paper.}, primarily through Function Calling (FC), has significantly expanded the boundaries of artificial intelligence~\citep{wang2025function,shen2024llm,wang2025empiricalstudyagentdevelope}. As agents are deployed in increasingly sophisticated scenarios—from enterprise workflow automation to intricate data analysis—the demand for high-quality, diverse, and complex training corpora has become the primary bottleneck limiting their advancement~\citep{liu2024apigen,zeng2025toolace}.
While recent works have made strides in synthesizing diverse datasets, existing data generation pipelines predominantly adhere to a paradigm of random sampling and shallow generation—randomly selecting API combinations from static tool pools and generating trajectories through basic user-assistant simulation~\citep{huang2025ttpa,prabhakar2025apigen}. 
Trajectories generated by those pipelines tend to be homogeneous and follow a “\textit{happy path}”, failing to capture the \textit{implicit logical dependencies} and \textit{multi-turn reasoning} required in hard real-world tasks.
For instance, real-world problems often necessitate bridging logical gaps, where the output of one tool serves as a latent precondition for another, or implicitly parameterizes the subsequent tool call. Current pipelines often miss these deep structural complexities, resulting in agents that are robust on simple queries but fragile when facing hard, logically-intertwined instructions~\citep{zhang2025looptool,yin2025magnet}.

To bridge this gap, we introduce \textbf{HardGen}, an automated agentic pipeline designed to synthesize challenging training samples for overcoming the performance bottleneck of tool-use agents. Unlike previous pipelines that focus primarily on broadening API coverage or enforcing syntactic correctness, HardGen prioritizes the logical complexity and difficulty of the generated trajectories. The core philosophy of HardGen is to \textit{learn from failure and evolve through feedback}. 
Specifically, instead of sampling from a static tool pool, HardGen establishes a dynamic API Graph derived explicitly from agent failure cases. This allows the pipeline to synthesize \textit{hard traces} that target the specific weaknesses of  models.
Next, we utilize these traces as conditional priors to guide the instantiation of modular, abstract advanced tools. These tools are then leveraged as stepping stones to formulate hard queries that necessitate multi-step reasoning and implicit dependency, moving beyond simple API pattern matching. 
At the end, the synergy of advanced tools and hard 
queries facilitates the generation of complex Chain-of-Thoughts (CoTs), with a closed-loop evaluation feedback mechanism steering the continuous refinement of the generation process, where reasoning correctness is further verified through a function call checking module. This robust, repeatable pipeline enables HardGen to construct trajectories of high diversity and complexity for tool-use agents.

\noindent\textbf{Generated Dataset.} 
With HardGen, we construct a comprehensive dataset containing 27,000 trajectories with 2,095 APIs from real environments (see \cref{sect:data}).
Our dataset encompasses a rich spectrum of complex interaction scenarios, particularly hard queries necessitating implicit logical bridging, where agents must autonomously infer tool dependencies and perform multi-step reasoning without explicit guidance. This large-scale, high-fidelity corpus is designed to overcome the complexity bottleneck in tool-use agent research, providing the community with a verifiable and rigorous foundation for training models capable of mastering deep logical dependencies.

\noindent\textbf{Remarkable Results.} 
We evaluate our pipeline by performing SFT or RL training on the Qwen3-4B~\cite{yang2025qwen3} model using the HardGen curated dataset. The resulting models, denoted as \textbf{HardGen-4B-SFT} and \textbf{HardGen-4B-RL}, are evaluated against several leading proprietary models on the challenging BFCLv3 \cite{patilberkeley} and other benchmarks.  Despite its compact \textbf{4B} scale, HardGen-4B-RL attains an overall accuracy of \textbf{79.14\%}, setting a new state-of-the-art record of its size (see \cref{tab:bfcl-overall}). 
For example, despite the massive disparity in model scale, HardGen-4B-RL achieves superior advantages over the latest strong competitors, surpassing GPT-5.2~\cite{openai2025gpt5.2} by \textbf{19.02\%}, DeepSeek-V3.2~\cite{liu2025deepseek} by \textbf{10.33\%} and Grok-4.1-Fast~\cite{xaigrok4.1} by \textbf{3.94\%} (see \cref{fig:perfo1}). Crucially, these gains generalize to \textbf{held-out} benchmark (BFCLv4~\cite{patilberkeley}), validating that the model learns robust agentic reasoning patterns rather than memorizing templates.

Our contributions are summarized as follows:
\begin{itemize}
    \item We propose {HardGen}, a novel pipeline targeting generating hard function-calling data for tool-use agents. 
    \item HardGen is compatible with a wide spectrum of APIs and models, facilitating the synthesis of tool-use datasets characterized by both high fidelity and complexity.
    \item Extensive experiments demonstrate that models trained on our generated data achieve superior performance, allowing a 4B parameter model to surpass powerful competitors. 
\end{itemize}

\begin{figure*}[thp]
    \centering
    \includegraphics[width=1.0\textwidth]{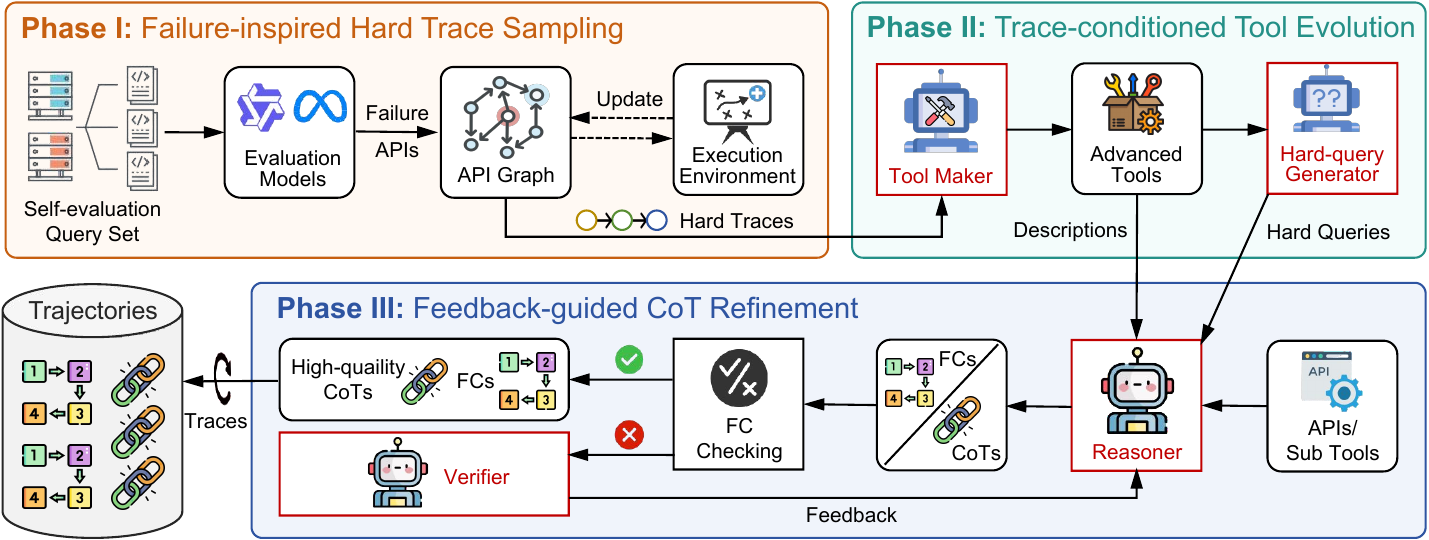}
    \caption{\textbf{Overview of the HardGen framework}.  The pipeline operates in three phases: \textbf{I) Failure-inspired Hard Trace Sampling} to identify error-prone tool dependencies and construct hard tool traces; \textbf{II) Trace-conditioned Tool Evolution} to synthesize advanced tools and hard queries based on the constructed hard traces; and \textbf{III) Feedback-guided CoT Refinement} to verify and optimize reasoning chains through a closed-loop mechanism.}
    \label{fig:pipeline}
\end{figure*}

\section{Related Work}

\noindent\textbf{Tool-use Ability of LLM.} 
Empowering LLMs to interact with external APIs is pivotal for the realization of autonomous agentic systems~\cite{wang2025function,FunReason}. By bridging the gap between static knowledge and dynamic execution, this paradigm enables models to interrogate databases, synthesize code, and manipulate digital interfaces~\cite{jimenez2023swe,mohammadjafari2024natural,OSWorld,gao2406simulating}. Yet, the transition from isolated function invocations to coherent, multi-turn tool orchestration presents a substantial barrier~\cite{guo2025deepseek,chen2025acebench,patilberkeley}. In these scenarios, agents are required to maintain state and resolve dependencies over extended interaction horizons.
The core bottleneck is the lack of large-scale, verifiable training corpora that faithfully encode realistic multi-turn interactions, including implicit preconditions, parameter couplings, and cross-turn logical dependencies~\cite{prabhakar2025apigen,xu2025toucan,yin2025magnet}. 

\noindent\textbf{Data Synthesis for Tool-use Training.} 
Current paradigms in tool-use data generation focus predominantly on expanding breadth and ensuring executability~\cite{yin2025magnet,acikgoz2025can}, typically via synthesis from fixed toolsets or multi-agent simulations~\cite{prabhakar2025apigen,yin2025magnet}. Although these methods scale dataset size~\cite{xu2025toucan}, they suffer from a critical limitation: the difficulty of the generated trajectories is often artificial—derived from explicit structural constraints—rather than reflecting the intrinsic reasoning hurdles and implicit dependencies characteristic of authentic agentic workflows~\cite{zeng2025toolace,lam2024closer,xu2025toucan}.
In contrast, our HardGen aims to generate complex, verifiable tool-use trajectories that deliver grounded reasoning for SFT and precise rewards for RL~\cite{qian2025toolrl,RLFC}, thereby driving superior performance in agentic tasks.

\section{The Proposed HardGen Pipeline}
\label{sec:methodology}


In this section, we present HardGen, an automatic agentic pipeline designed to generate hard tool-use training samples with verifiable reasoning.  

\subsection{Overview}

As the overview illustrated in \cref{fig:pipeline} and notations shown in \cref{tab:coresets}, HardGen consists of three coordinated phases: 
%
\noindent \textbf{First}, we identify error-prone function calls via model evaluation on a self-evaluation query set. These “failure APIs” are then structured into a dynamic API Graph, which is iteratively updated through interactions with an execution environment to capture complex tool dependencies.
\noindent \textbf{Second}, we sequentially feed the generated tool traces from the API Graph into a \textit{Tool Maker} and a \textit{Hard-query Generator}, which evolves simple functions into advanced tools and synthesize corresponding high-complexity queries.
\noindent \textbf{Third}, we employ a \textit{Reasoner}-\textit{Verifier} loop to execute these hard queries, using environment feedback to rigorously filter and refine function calls and CoTs.
By repeating this three-phase process, our HardGen generates reliable and complex multi-turn trajectories, which contain hard queries, primitive tools, verified CoTs and function calls.

\begin{table}[tbp]
\centering
\small
\begin{tabular}{lp{4.3cm}}
\toprule
\textbf{Symbol} & \textbf{Concept} \\
\midrule
$\mathcal{S}$ & Environment Simulation Space \\
$\mathcal{T}$ & Failure API Set \\
$\mathcal{G} = (\mathcal{T}, \mathcal{D}, \mathcal{P})$ & API Graph (Failure API Set $\mathcal{T}$, Dependencies $\mathcal{D}$, Parameters $\mathcal{P}$) \\
$A_T, A_Q, A_R, A_V$ & Agents: Tool Maker, Hard-query Generator, Reasoner, Verifier \\
$T_a \in \mathcal{T}$ & Selected Tool \\
$Q_{hard}$ &Hard Query\\
$T_{adv}$ &Advanced Tool\\
$M$ & Number of Tool Calls Per Trace \\
$N$ & Number of Turns Per Trajectory  \\
\bottomrule
\end{tabular}
\caption{\textbf{Core notations}. Summary of symbols and definitions used in this work.}
\label{tab:coresets}
\end{table}



\subsection{Phase I: Failure-inspired Hard Trace Sampling}
\label{sec:phase1}
Rather than uniformly sampling tool combinations, this phase leverages a failure-driven self-evaluation to surface tools and dependencies that the current model struggles with. The resulting API Graph serves as a structured representation of failure-prone tools and their latent interaction patterns, from which a series of hard and valuable tool traces can be sampled.

\paragraph{Failure-driven Self-evaluation and Graph Construction.}


We deploy a large-scale API environment comprising 2,095 tools to perform model self-evaluation. For each tool, HardGen generates an execution trace and constructs a corresponding hard query, forming a Self-evaluation Query Set used to evaluate Qwen3-4B~\cite{yang2025qwen3} and Llama-3.2-3B-Instruct~\cite{dubey2024llama}. A tool is designated as challenging if both models produce incorrect execution results during inference. Through this process, we identify 1,204 challenging tools, which are incrementally incorporated into both the Failure API Set $\mathcal{T}$ and API Graph $\mathcal{G}$, enabling systematic capture of challenging tools and their dependencies. The graph construction and its update are shown in the \textbf{Appendix}~\ref{app:graph}. Despite performing evaluation and construction solely on Qwen3 models, our approach demonstrates robust performance across heterogeneous architectures including Llama-3 (see \cref{tab:llamaModels}) and Qwen2.5 (see \cref{tab:Qwen2.5}), validating the strong generalization capability of the HardGen synthesized data.



\paragraph{Legality-constrained Sampling.} 
To guarantee execution validity, tool selection is subject to a strict dependency constraint: a tool $T_a$ is callable if and only if all its prerequisite dependency tools ($\mathcal{D}_{T_a}$) have been executed previously, denoted by $\mathcal{T}_{\text{called}}$. This legality condition is formalized as
\begin{equation}
I(T_a, \mathcal{T}_{\text{called}}) = \mathbf{1}_{\{\mathcal{D}_{T_a} \subseteq \mathcal{T}_{\text{called}}\}},
\end{equation}
where $\mathbf{1}_{\{\cdot\}}$ is the indicator function.


\paragraph{Hard Trace Sampling.}
To reach the selected target tool $T_a$ while satisfying its prerequisite constraints, we design a \emph{Sampler} that explicitly biases trace construction toward $T_a$. Specifically, the Sampler employs a greedy heuristic over the set of legal tools $\mathcal{T}_{\text{legal}}$, prioritizing tools that minimize the graph distance $\text{dist}(\cdot,\cdot)$ to $T_a$. Here, $\text{dist}(T_k, T_a)$ denotes the length of the shortest path from $T_k$ to $T_a$ in the API graph $\mathcal{G}$. The resulting sampling policy is formally defined as:
\begin{equation}
\label{eq:tool_sampling}
\begin{aligned}
T_s
&= \operatorname{Sampler}(T_a, \mathcal{T}_{\text{called}}) \\
&=
\left\{
\begin{array}{@{}l@{\quad}l@{}}
\operatorname{rand}(\mathcal{T}_{\text{legal}}),
& T_a \in \mathcal{T}_{\text{called}}, \\[4pt]
T_a,
& \hspace{-8em} I(T_a, \mathcal{T}_{\text{called}})=1 \land T_a \notin \mathcal{T}_{\text{called}}, \\[4pt]
\displaystyle
\operatorname*{arg\,min}_{T_k}
\operatorname{dist}(T_k, T_a),
& \text{otherwise}.
\end{array}
\right.
\end{aligned}
\end{equation}

The sampled tool $T_s$ is executed with parameters $P_s$ as a call $C_s = (T_s, P_s)$, producing environment feedback $E_s$ and updating the system state $\mathcal{S}$, the set of executed tools $\mathcal{T}_{\text{called}}$, and the dynamic API graph $\mathcal{G}$. The resulting execution sequence is recorded as an executable hard trace $\Gamma_i = (C_1,E_1,C_2, E_2, \dots,C_M, E_M)$, which serves as a failure-aware prior for subsequent phases. 





\subsection{Phase II: Trace-conditioned Tool Evolution}
\label{sec:phase2}



The hard trace $\Gamma_i$ generated in Phase I guarantees correct tool execution and faithfully captures the underlying operational sequence. However, a central challenge in training robust tool-use agents lies in enabling \emph{implicit logical bridging}—the ability to autonomously infer the necessary intermediate steps when faced with hard queries that do not explicitly specify the full tool chain. Directly generating queries from execution traces, where all intermediate steps are explicitly enumerated, fails to cultivate this capability, as models can simply rely on pattern matching rather than learning to bridge logical gaps. To address this challenge, Phase II reinterprets the trace via a two-stage evolution process: it first abstracts the multi-step execution into a unified high-level operation, and then constructs a challenging hard query that explicitly requires this abstraction.


\paragraph{Advanced Tool Construction.}
Given  $\Gamma_i$, the \emph{Tool Maker} ($A_T$) synthesizes a unified high-level operation abstraction, denoted as the advanced tool $T_{\text{adv}}$:
\begin{equation}
T_{\text{adv}} = A_T\left(\Gamma_i\right),
\end{equation}
where $A_T$ abstracts the multi-step execution trace into a high-level operation that encapsulates the collective functionality and interdependencies of all tool calls within $\Gamma_i$.

\paragraph{Hard Query Generation.}
Conditioned on the synthesized advanced tool $T_{\text{adv}}$, the \emph{Hard-query Generator} ($A_Q$) generates a challenging query $Q_{\text{hard}}$ that explicitly requires the use of $T_{\text{adv}}$ for resolution.  
Formally, the process is expressed as:
\begin{equation}
Q_{\text{hard}} = A_Q\left(T_{\text{adv}}\right).
\end{equation}

\subsection{Phase III: Feedback-guided CoT Refinement}
\label{sec:phase3}

While the hard query $Q_{\text{hard}}$ constructed in Phase II effectively challenges the model with implicit logical dependencies, it also amplifies the difficulty of generating correct and coherent Chain-of-Thought (CoT) reasoning. To mitigate this issue, Phase III implements an iterative refinement mechanism that progressively corrects flawed reasoning via feedback-driven prompting, guiding the CoT from erroneous states toward correct solutions.

\paragraph{Reasoning with Hint.}
Given the hard query $Q_{\text{hard}}$, the \emph{Reasoner} $A_R$ attempts to generate the correct function call with the hint from the description of the advanced tool $T_{\text{adv}}$. The initial prompt, denoted as $\text{Prompt}^{(1)}_i$, is constructed from the hard query, the primitive tool set used to define the advanced tool, and the advanced tool description.


\paragraph{Error Diagnosis.} 
When the current attempt fails (\textit{i.e.}, $\text{FC}^{(k)}_i \neq C_i$), the \emph{Verifier} $A_V$ analyzes the discrepancy between the incorrect function call and the ground truth. The Verifier identifies the specific error and generates corrective hint feedback $\text{Error}^{(k)}_i$ that guides correction without exposing the answer.

\paragraph{CoT Refinement.} 
For step $i \in \{1, \ldots, M\}$ at its $k$-th attempt , we incorporate the corrective hint $\text{Error}^{(k)}_i$ into the current prompt through concatenation: 
\begin{equation}
    \text{Prompt}^{(k+1)}_i = \text{Concat}(\text{Prompt}^{(k)}_i, \text{Error}^{(k)}_i).
\end{equation}
With this augmented prompt, $A_R$ is re-invoked to produce a refined solution, yielding an updated reasoning process $\text{CoT}^{(k+1)}_i$ and function call $\text{FC}_i^{(k+1)}$. This iterative refinement continues until either the function call is correct or the maximum number of attempts $K{\max}$ is reached.

Upon successfully generating the correct function call $C_i$ at step $i$ on the $k$-th attempt, the environment execution feedback $E_i$ is incorporated into the prompt for the subsequent $i+1$ step via concatenation: $\text{Prompt}^{(1)}_{i+1} = \text{Concat}(\text{Prompt}^{(k)}_i, C_i, E_i)$. This design ensures that subsequent reasoning is explicitly conditioned on the complete execution history accumulated up to the current step. A trace is retained only if all $M$ function calls are correctly generated across the entire sequence.

\subsection{Generated Dataset}
\label{sect:data}
\begin{figure}[t]
    \centering
    \begingroup\setlength{\parskip}{0pt} 
    \begin{minipage}{\linewidth}
        \centering
        \begin{minipage}{0.58\linewidth}
            \centering
            \includegraphics[width=\linewidth,trim=0 12pt 0 0,clip]{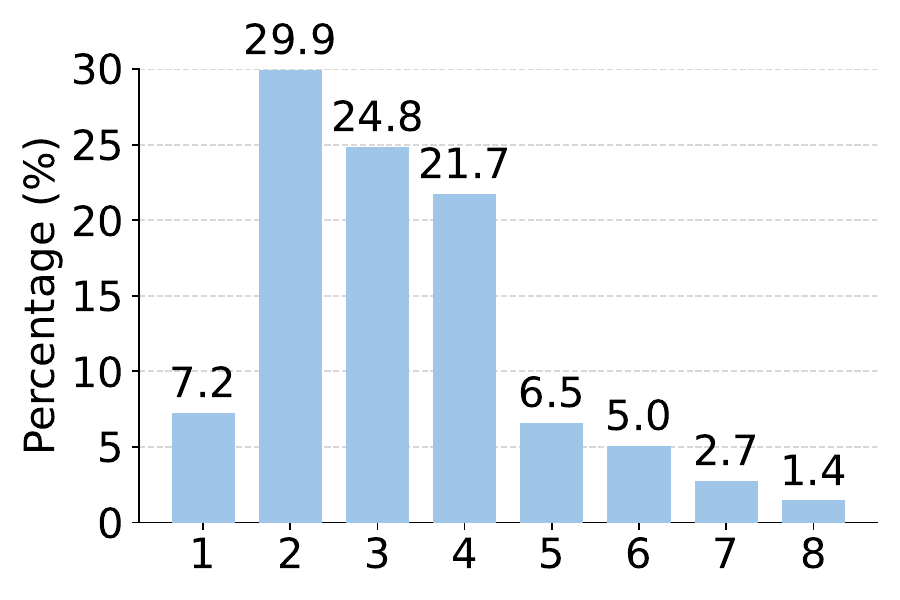}
        \end{minipage}%
        \hfill
        \begin{minipage}{0.42\linewidth}
            \centering
            \scriptsize
            \begin{tabular}{lc}
                \toprule
                \textbf{Metric} & \textbf{Value} \\
                \midrule
                Total Trajectories& 27,000\\
                Max. Tool Calls & 8\\
                Min. Tool Calls & 1\\
                Avg. Tool Calls & 3.21\\
                Min. Turns & 1 \\
                Max. Turns & 8 \\
                Avg. Turns & 3.32 \\
                \bottomrule
            \end{tabular}
        \end{minipage}
    \end{minipage}
    \begin{minipage}{\linewidth}
        \centering
        \includegraphics[width=1\linewidth]{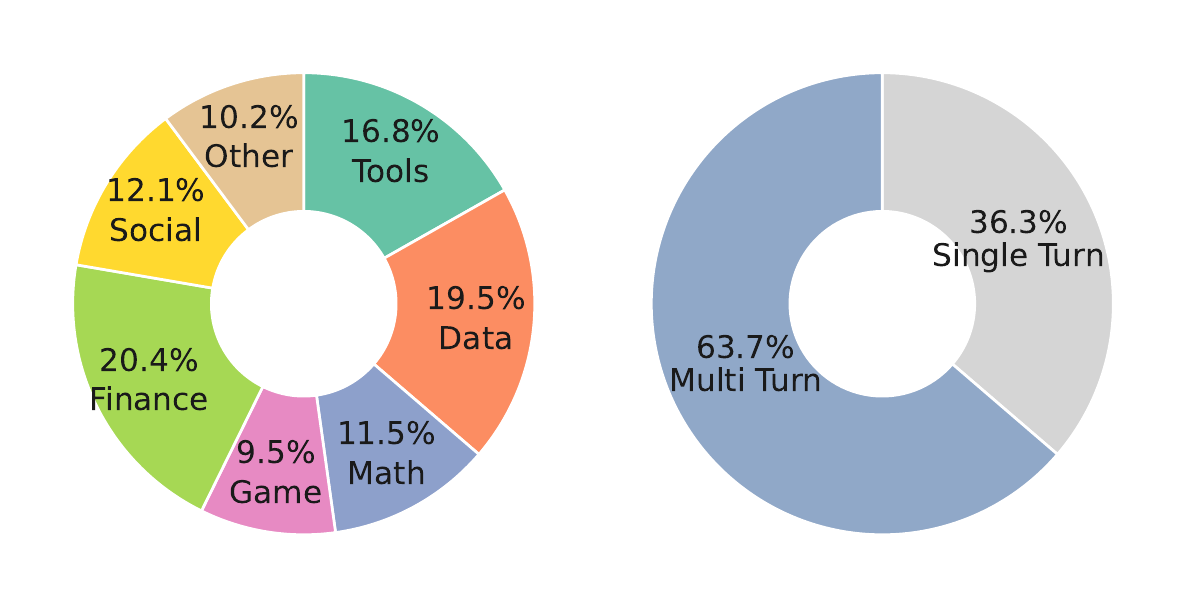}
    \end{minipage}
    \endgroup
    \caption{\textbf{Statistics and distribution of the generated dataset}.  \textbf{(Top)} Histogram of tool calls per trajectory (left) and key dataset metrics (right). \textbf{(Bottom)} Distribution of API domains (left), the proportion of single-turn and multi-turn trajectories (right).}
    \label{fig:dataset}
\end{figure}


\begin{table*}[!t]
\footnotesize
\centering
\resizebox{\textwidth}{!}{
\begin{tabular}{lcccccccccc}
\toprule
& & \multicolumn{3}{c}{\textbf{Single-Turn}} & \multicolumn{5}{c}{\textbf{Multi-Turn}} & \\
\cmidrule(lr){3-5}\cmidrule(lr){6-10}
\textbf{Model}
& \begin{tabular}[c]{@{}c@{}}\textit{Parameter}\\ \textit{Counts}\end{tabular}
& \textit{Non-Live}
& \textit{Live}
&\begin{tabular}[c]{@{}c@{}}\textit{Subset}\\ \textit{Overall}\end{tabular}
& \textit{Base}
& \begin{tabular}[c]{@{}c@{}}\textit{Miss}\\ \textit{Func}\end{tabular}
& \begin{tabular}[c]{@{}c@{}}\textit{Miss}\\ \textit{Param}\end{tabular}
& \begin{tabular}[c]{@{}c@{}}\textit{Long}\\ \textit{Context}\end{tabular}
& \begin{tabular}[c]{@{}c@{}}\textit{Subset}\\ \textit{Overall}\end{tabular}
&\textbf{Overall} \\
\midrule
\rowcolor{gray!15}
\multicolumn{11}{c}{\emph{\textbf{Closed-source}}} \\
Claude-Opus-4-5-20251101 & - &88.58 &79.79 &84.19 &81.00 &64.00 &58.00 &70.50 &68.38 &78.92\\
Claude-Sonnet-4-5-20250929 & - &88.65 &81.13 &84.89 &69.00 &65.00 &52.50 &59.00 &61.38 &77.05\\
Claude-Haiku-4-5-20251001 & - &86.50 &78.68 &82.59 &63.50 &42.50 &52.50 &56.00 &53.63 &72.93\\
Gemini-3-Pro-Preview & - &90.65 &83.12 &86.89 &64.50 &60.00 &54.50 &64.00 &60.75 &78.17\\
Gemini-2.5-Flash & - &84.96 &74.39 &79.68 &41.50 &36.00 &32.00 &35.50 &36.25 &65.20\\
Grok-4-1-fast-reasoning & - &88.27 &78.46 &83.37 &70.50 &59.50 &43.00 &62.50 &58.88 &75.20\\
Grok-4-1-fast-non-reasoning & - &88.13 &77.94 &83.04 &58.00 &39.50 &37.50 &52.00 &46.75 &70.94\\
GPT-5.2-2025-12-11 & - &81.85 &70.39 &76.12 &36.50 &18.00 &27.50 &30.50 &28.13 &60.12\\
GPT-4o-2024-11-20 & - &83.88 &70.54 &77.21 &55.50 &34.50 &29.00 &51.00 &42.50 &65.64\\
\midrule
\rowcolor{gray!15}
\multicolumn{11}{c}{\emph{\textbf{Open-source}}} \\
Kimi-K2-Instruct & 1043B &81.60 &78.68 &80.14 &62.00 &41.00 &44.50 &55.00 &50.63 &70.30\\
DeepSeek-V3.2-Exp & 671B &85.52 &76.02 &80.77 &55.00 &49.00 &27.00 &48.50 &44.88 &68.81\\
Llama-4-Maverick & 400B &88.65 &73.65 &81.15 &27.00 &22.00 &14.00 &18.00 &20.25 &60.85\\
GLM-4.6 & 355B &87.56 &80.90 &84.23 &74.50 &68.00 &63.00 &66.50 &68.00 &78.82\\
Qwen3-235B-A22B-Instruct & 235B &90.33 &78.68 &84.51 &54.00 &42.50 &31.50 &50.50 &44.63 &71.21\\
Qwen3-32B & 32B &88.77 &82.01 &85.39 &56.00 &52.50 &40.00 &43.00 &47.88 &72.88\\
Qwen3-30B-A3B-Thinking & 30B &85.77 &77.94 &81.86 &43.50 &10.50 &25.00 &41.00 &30.00 &64.57\\
ToolACE-2-8B & 8B &87.10 &77.42 &82.26 &49.00 &28.00 &30.50 &46.00 &38.38 &67.63\\
ToolACE-MT & 8B &84.94 &71.52 &78.23 &57.50 &31.50 &34.00 &38.00 &40.25 &65.57\\
Nanbeige4-3B-Thinking-2511&3B&81.58	&79.42&	80.50		&58.50	&54.00	&45.00	&47.00	&51.12 &70.71\\
xLAM-2-3b-fc-r & 3B &82.96 &62.92 &72.94 &71.50 &59.00 &57.50 &45.50 &58.38 &68.09\\
\midrule
Qwen3-4B (Base Model) & 4B &87.88 &76.39 &82.14 &26.50 &21.00 &15.50 &25.50 &22.13 &62.13\\
\rowcolor{green!10}\textbf{HardGen-4B-SFT} & 4B &89.03 &82.42 &85.73 &55.20 &49.10 &41.40 &52.90 &49.65 &73.70\\
\rowcolor{green!10}{$\Delta$} & & \textcolor{red}{+1.15}&\textcolor{red}{+6.03}&\textcolor{red}{+3.59} & \textcolor{red}{+28.70}& \textcolor{red}{+28.10}& \textcolor{red}{+25.90} & \textcolor{red}{+27.40}&\textcolor{red}{+27.52}&\textcolor{red}{+11.57}\\
\rowcolor[HTML]{EAE8FD}
\textbf{HardGen-4B-RL} & 4B &90.73 &83.55 &87.14 &68.50 &64.50 &50.50 &69.00 &63.13 &\textbf{79.14}\\
\rowcolor[HTML]{EAE8FD}
{$\Delta$} & & \textcolor{red}{+2.85}&\textcolor{red}{+7.16}&\textcolor{red}{+5.00} & \textcolor{red}{+42.00} & \textcolor{red}{+43.50}& \textcolor{red}{+35.00}& \textcolor{red}{+43.50}&\textcolor{red}{+41.00}&\textcolor{red}{+17.01}\\
\bottomrule
\end{tabular}}
\caption{\label{tab:bfcl-overall} \textbf{Performance on BFCLv3} (last updated on 2025-12-16). All metrics are calculated using the official script and reported in terms of Accuracy (\%).}
\end{table*}

We deploy HardGen on a large-scale real-world API environment comprising 2,095 tools, of which 1,204 are identified as Failure Tools through systematic Self-evaluation. Training trajectories are synthesized using the HardGen pipeline, with Qwen3-30B-A3B-Thinking~\cite{yang2025qwen3} serving as the backbone of the four agents. The system employs a maximum attempt limit of $K_{\max}=3$, which we find offers a favorable balance between trajectory quality and computational cost. As shown in \cref{fig:dataset}, the resulting dataset comprises 27,000 high-quality trajectories that exhibit substantial diversity across multiple dimensions. First, in terms of task complexity, trajectories span 1 to 8 tool calls (with an average of 3.21), and 62.1\% of them contain three or more tool calls within a single trace. Second, regarding API coverage, the dataset spans diverse domains including Finance (20.4\%), Game (9.5\%), Tools (16.8\%), and Data (19.5\%), ensuring broad applicability. Third, the trajectories capture rich reasoning patterns, from straightforward single-turn executions (36.3\%) to complex multi-turn scenarios (63.7\%) requiring sophisticated inference capabilities. A supplementary trajectory case and the prompts of corresponding agents are shown in \textbf{Appendix}~\ref{app:additional_results} and \textbf{Appendix}~\ref{appendix:prompt}.

\section{Experiments}
\label{sec:experiments}

In this section, we evaluate our proposed pipeline by performing Supervised Fine-Tuning (\textbf{SFT}) and Reinforcement Learning (\textbf{RL}) on several baseline models using the HardGen curated dataset. 

\subsection{Experiment Setup}
\label{subsec:exp_setup}

\noindent\textbf{Baseline Models.} We employ Qwen3-4B as our main baseline model~\cite{yang2025qwen3}. To validate the generalizability of our pipeline, we additionally conduct experiment with Qwen2.5-7B-Instruct~\cite{team2024qwen2}, Llama-3-3B/8B-Instruct~\cite{dubey2024llama}, and Qwen3-0.6B/1.7B. We split the 27,000 trajectories at each assistant response and train the model to generate only the current-turn response. We perform SFT with LLaMA-Factory~\cite{zheng2024llamafactory} and RL with Verl~\cite{sheng2024hybridflow}. 
Implementations are detailed in \textbf{Appendix}~\ref{app:trainingdetails}.

\noindent\textbf{Benchmarks.} We evaluate our models on BFCLv3~\cite{patilberkeley}, API-Bank~\cite{li2023api}, and ACEBench~\cite{chen2025acebench}, covering both single-turn and multi-turn tool-calling scenarios (see \textbf{Appendix}~\ref{app:trainingdetails}). To evaluate out-of-distribution generalization and assess how the proposed training data impacts specific agentic capabilities, we include BFCLv4 as a \textbf{held-out} benchmark~\cite{patilberkeley}. BFCLv4 offers a comprehensive evaluation of agentic behaviors in Web Search and Memory scenarios, with details provided in \textbf{Appendix}~\ref{appendix:bfclv4}.

\subsection{Experimental Results}
\label{subsec:main_results}

\noindent\textbf{Results on BFCLv3.}  As shown in \cref{tab:bfcl-overall}, on the BFCLv3 benchmark~\cite{patilberkeley}, models trained with HardGen  yield notable improvements on Qwen3-4B, raising the multi-turn score from 22.13 to 49.65 (\textbf{+27.52}) after SFT and to 63.13  (\textbf{+41.01}) after RL. Despite its 4B parameter size, the HardGen RL-trained model surpasses strong open-source models (\textit{e.g.}, Kimi-K2-Inst~\cite{team2025kimi}, DeepSeek-V3.2~\cite{liu2025deepseek}) and leading closed-source models (e.g., GPT-5.2~\cite{openai2025gpt5.2}, Gemini-3-Pro~\cite{gemini3pro}, Claude-Opus-4.5~\cite{claudeopus4.5}), setting a new state-of-the-art record of its size. In addition, models trained with HardGen demonstrate balanced performance across all sub-metrics, indicating strong generalization and stability. A particularly compelling result is that both HardGen-4B-SFT and HardGen-4B-RL outperform Qwen3-30B-A3B-Thinking~\cite{yang2025qwen3} by a substantial margin across all sub-metrics. This result validates that our failure-driven sampling, trace-conditioned tool evolution, and feedback-guided refinement jointly form a virtuous cycle of capability amplification, effectively transcending the inherent limitations of the generator model. To further assess the generalizability of HardGen-synthesized data, we report results on the Llama model family in \textbf{Appendix}~\ref{appendix:othermodels}, demonstrating consistent improvements across diverse model architectures.

\begin{figure*}[t]
        \centering
        \includegraphics[width=\linewidth]{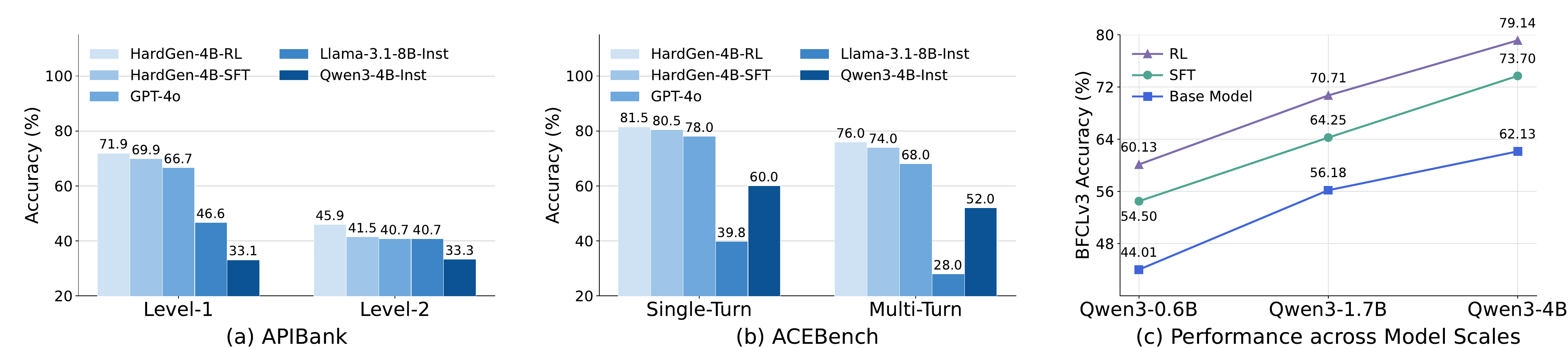}
    \caption{\textbf{Additional evaluations}. \textbf{(a) (b)} Comparison of different models on APIBank and ACEBench. \textbf{(c)} Scaling trends of HardGen-SFT and HardGen-RL on BFCLv3 across model parameters.}
    \label{fig:toolandturn}
\end{figure*}

\begin{table}[tbp]
\centering
\small
\resizebox{0.49\textwidth}{!}{
\begin{tabular}{lccc}
\toprule
\textbf{Model} & \textit{Single-Turn} & \textit{Multi-Turn} & \textbf{{Overall}}\\
\midrule
Qwen2.5-7B-Instruct (Base Model) & 77.30 & 7.62& 54.07\\
\midrule
TOUCAN~\cite{xu2025toucan} &76.51 &22.62&58.55\\
MAGNET~\cite{yin2025magnet} &80.49&21.12&60.70 \\
ToolACE-MT~\cite{zeng2025toolace}&80.51 &27.57 &62.86\\
\midrule
\rowcolor[HTML]{EAE8FD}+ \textbf{HardGen-SFT} &\textbf{83.99}&\textbf{40.75}&\textbf{69.58}\\
\rowcolor[HTML]{EAE8FD}{$\Delta$} & \textcolor{red}{+6.69}         & \textcolor{red}{+33.13}      & \textcolor{red}{+15.51}\\
\bottomrule
\end{tabular}}
\caption{\textbf{Comparison with state-of-the-art data synthesis pipelines on BFCLv3}. The baseline model is Qwen2.5-7B-Instruct.}
\label{tab:Qwen2.5}
\end{table}

\noindent\textbf{Results on APIBank and ACEBench.}  
\cref{fig:toolandturn}\textbf{(a)} and \cref{fig:toolandturn}\textbf{(b)} present the performance of our models on two additional benchmarks, APIBank and ACEBench. 
On APIBank~\cite{li2023api}, our models achieve top-tier Level-1 accuracies of 71.9 and 69.9, clearly outperforming GPT-4o~\cite{hurst2024gpt}, which attains 66.7. For the more challenging Level-2 tasks, our models continue to demonstrate strong performance, yielding improvements of \textbf{+12.6} and \textbf{+8.2} percentage points over the base model (33.3), respectively. Evaluation on ACEBench~\cite{chen2025acebench} further confirms robust generalization under both training configurations. On the single-turn subset, our models reach accuracies of 81.5 and 80.5, surpassing both GPT-4o (78.0) and Llama-3.1-8B-Instruct (39.8) by substantial margins. This advantage is more pronounced in the multi-turn setting, where our models achieve scores of 76.0 and 74.0, exceeding Llama-3.1-8B-Instruct (28.0) by \textbf{48.0} and \textbf{46.0} percentage points, respectively. Overall, these results provide compelling evidence that models trained on HardGen synthesized dataset exhibit strong and consistent tool-use capabilities across diverse benchmarks and interaction settings.


\noindent\textbf{Comparison with State-of-the-art.} To further demonstrate the efficacy of HardGen, we compare it against state-of-the-art data synthesis pipelines using Qwen2.5-7B-Instruct under an SFT-only setting, as reported in \cref{tab:Qwen2.5}. While prior methods such as MAGNET~\cite{yin2025magnet}, TOUCAN~\cite{xu2025toucan}, and ToolACE-MT~\cite{zeng2025toolace} offer incremental gains over the base model, HardGen establishes a clear and consistent performance advantage. Notably, our method secures a score of 83.99 (\textbf{+6.69}) in single-turn tasks and delivers a striking 40.75 (\textbf{+33.13}) in multi-turn interactions. This substantial margin validates HardGen as a more robust and effective data generation strategy for complex tool-use scenarios.


\subsection{Ablation Study}
In this section, we present ablative results to further scrutinize our proposed HardGen pipeline.

\begin{table}[t]
    \centering
    \resizebox{0.48\textwidth}{!}{
    \begin{tabular}{ccccc}
        \toprule
        \multirow{2}{*}{\centering \textbf{Hard Query}}
        & \multicolumn{2}{c}{\textit{Qwen3-4B}}
        & \multicolumn{2}{c}{\textit{Llama-3-3B}} \\
        \cmidrule(lr){2-3} \cmidrule(lr){4-5}
        & \textit{Single-Turn} & \textit{Multi-Turn}
        & \textit{Single-Turn} & \textit{Multi-Turn} \\
        \midrule
        \XSolidBrush & 84.17 & 54.38 & 79.08 & 32.23 \\
        \Checkmark  & \textbf{87.14} & \textbf{63.13} & \textbf{84.05} & \textbf{40.13} \\
        \bottomrule
    \end{tabular}}
        \caption{\textbf{Impact of hard queries}. Both Qwen3-4B and Llama-3-3B are trained with RL using data with or without hard queries.}
    \label{tab:performance}
\end{table}

\noindent\textbf{Does HardGen's effectiveness scale with model size?} To investigate the scalability of our approach, we evaluate the performance of the HardGen synthesized dataset across base models ranging from 0.6B to 4B parameters on the BFCLv3 benchmark. As shown in \cref{fig:toolandturn}\textbf{(c)}, model performance consistently improves with increasing scale under both training paradigms. Specifically, the Qwen3-0.6B model registers a gain of \textbf{+13.74} points (rising from 34.93 to 48.67) via RL, whereas Qwen3-4B realizes a more substantial improvement of \textbf{+16.93} points (60.21 to 77.14). Furthermore, RL consistently outperforms SFT across all model sizes, with the performance gap widening as the base model scales up. This trend indicates that HardGen-synthesized data is inherently well aligned with RL, enabling stronger base models—particularly at larger scales—to more effectively leverage RL for improving tool-use capabilities.

\noindent\textbf{How much do hard queries affect the quality of the synthesized data?} To assess the impact of the constructed hard queries on data quality, we perform ablation studies by conducting RL training on Qwen3-4B and Llama-3-3B. As shown in \cref{tab:performance}, incorporating hard queries leads to consistent performance improvements across both model architectures and evaluation tasks. Specifically, for Qwen3-4B, training with hard queries yields notable gains of \textbf{+8.75} points in the multi-turn subset (from 54.38 to 63.13) and \textbf{+2.97} points in the single-turn subset (from 84.17 to 87.14). Similarly, Llama-3-3B benefits from hard queries, achieving improvements of \textbf{+7.90} and \textbf{+4.97} points in the multi-turn and single-turn subsets, respectively.
The consistent gains observed across architectures and evaluation tasks demonstrate that the constructed hard queries substantially enhance the quality of the synthesized data, leading to stronger tool-use capabilities, particularly in more challenging multi-turn scenarios, where the improvements are most pronounced.

\begin{table}[t]
\centering

\resizebox{0.425\textwidth}{!}{
\begin{tabular}{lcc}
\toprule
\textbf{Model} & \textit{w/o $T_{adv}$} & \textit{w/ $T_{adv}$} \\
\midrule
\makecell[l]{Qwen3-30B-A3B-Thinking}  & 177 & \textbf{223} \\
\makecell[l]{Qwen3-235B-A22B-Thinking}  & 159 & \textbf{241} \\
\makecell[l]{DeepSeek-V3.1 (671B)} & 125& \textbf{275}  \\
\bottomrule
\end{tabular}}
\caption{\textbf{Impact of the constructed advanced tools $T_{adv}$ across different model scales}. The values denoted the number of instances deemed more difficult by GPT-4o from a pool of 400 generated queries.}
\label{tab:advtool}
\end{table}


\noindent\textbf{Do advanced tools really help with making logical bridging?} To evaluate the contribution of the constructed advanced tools $T_{adv}$ to logical jump bridging, we conduct a controlled comparison between model variants with and without $T_{adv}$ under identical configurations. Each model is evaluated on 400 query-synthesis instances, where every instance produces two queries: one that leverages $T_{adv}$ for logical bridging and one that does not. GPT-4o is employed as an automated judge to determine which variant yields a more challenging query, \textit{i.e.}, one involving more difficult logical jumps. As shown in \cref{tab:advtool}, incorporating $T_{\mathrm{adv}}$ consistently increases the proportion of challenging queries across all model scales, with the number of such instances rising from 223 to 275 as model capacity increases from 30B to 671B. This trend indicates that $T_{adv}$ effectively facilitates logical bridging, which is more pronounced with increasing model capacity. Since the subjective nature of \emph{difficulty} judgments, we further complement the automated evaluation with a human annotation study in \textbf{Appendix}~\ref{appendix:manualann}, using the same rubric for assessing logical-jump difficulty. The results show a high level of agreement between the automated judge and human annotators, supporting the conclusion that $T_{adv}$ reliably increases implicit logical-bridging difficulty.




\begin{table}[t]
    \centering
    \small
    \resizebox{0.48\textwidth}{!}{
\begin{tabular}{lccc}
\toprule
\textbf{Model} & \textit{w/o Feedback} & \textit{w/ Feedback} &{$\Delta$}\\
\midrule
\makecell[l]{Qwen3-30B-\\A3B-Thinking} & 77.80\% & 90.14\% &\textcolor{red}{+12.34\%} \\
\midrule
\makecell[l]{Qwen3-235B-\\A22B-Thinking} & 82.57\% & 92.78\%&\textcolor{red}{+10.21\%} \\
\midrule
\makecell[l]{DeepSeek-\\V3.1 (671B)} & 83.98\% &95.60\%&\textcolor{red}{+11.62\%} \\
\bottomrule
\end{tabular}}
\caption{\textbf{Impact of feedback-guided CoT refinement}. 
The three baseline models are trained with RL using data with or without Verifier  feedback.
}
\label{tab:CoT}
\end{table}


\noindent\textbf{How significantly does feedback-guided CoT refinement boost reasoning?} To quantify the impact of our feedback-guided CoT refinement, we compare model variants with and without Verifier feedback. 
Concretely, we set $K_{\max}$ to 3 as the maximum number of refinement iterations. As reported in \cref{tab:CoT}, incorporating the guided refinement loop results in substantial accuracy gains of \textbf{+12.34} on the 30B model, \textbf{+10.21} on the 235B model, and \textbf{+11.62} on the 671B model. These consistent improvements—each exceeding 10 more percentage points across all evaluated model scales—highlight the critical role of feedback-guided refinement in steering model reasoning toward correct solutions, establishing it as an essential component of our framework. A detailed analysis of the refinement dynamics and the choice of $K_{\max}$ is provided in \textbf{Appendix}~\ref{appendix:data}.

\section{Conclusion}
\label{sec:conclusion}

In this work, we introduce HardGen, an automatic agentic pipeline designed to generate challenging tool-use training samples with verifiable reasoning. HardGen adopts a failure-driven approach, producing training samples that capture the implicit logical dependencies and multi-step reasoning characteristic of real-world tasks. Extensive experiments demonstrate that a 4B parameter model trained with our curated dataset achieves state-of-the-art performance on BFCLv3 for its scale, surpassing leading proprietary models. HardGen exhibits strong generalization across model architectures, parameter scales, and held-out benchmarks, paving the way for developing more robust models capable of complex tool-use and agentic ability.


\section*{Limitation}

While HardGen demonstrates strong performance across multiple benchmarks and model architectures, several limitations merit consideration. Although our evaluation covers 2,095 tools across diverse domains, the extent to which HardGen generalizes to entirely new API ecosystems or specialized domains remains to be fully explored. 
Our approach requires executable environments for verification, which may not be feasible for proprietary APIs or tools with complex external dependencies. 
The abstraction quality of advanced tools relies on the Tool Maker's ability to correctly identify high-level operations from primitive tool sequences, which may occasionally produce suboptimal abstractions for highly irregular or domain-specific tool chains.

\bibliography{custom}

\appendix

\section{Data}
\label{appendix:data}



\begin{figure}[h]
    \centering
    \includegraphics[width=1.0\linewidth]{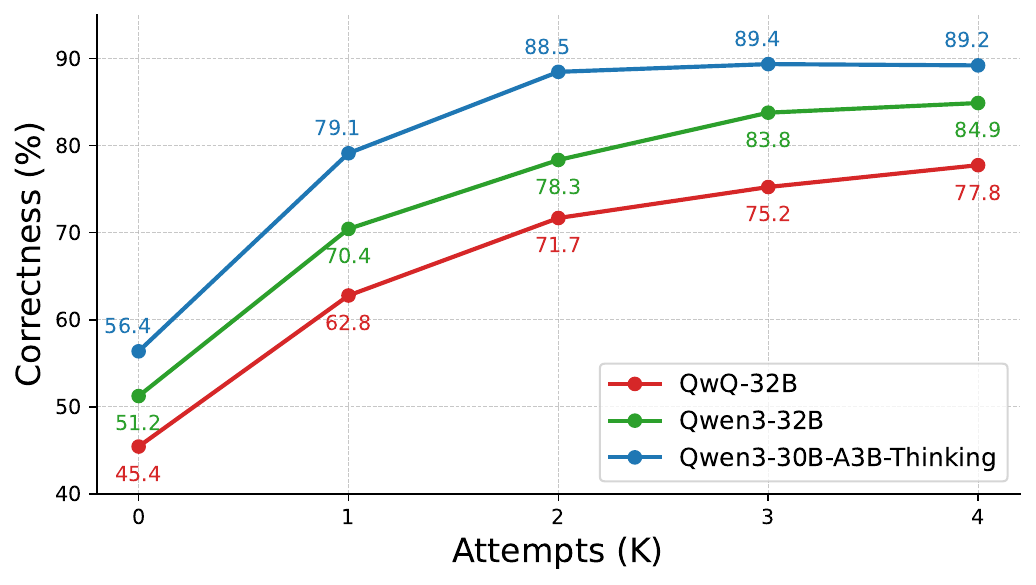}
    \caption{\textbf{Analysis of generator model selection}. Correctness rates of function call synthesis at different numbers of attempts ($K$) for three candidate generator models.}
    \label{fig:kattempt}
\end{figure}

\noindent\textbf{Model Selection and Number of Attempts}. To ensure computational efficiency for large-scale synthesis, we restrict candidates models around 30B parameter scale and evaluate three strong generators—QwQ-32B~\cite{team2025qwq}, Qwen3-32B~\cite{yang2025qwen3}, and Qwen3-30B-A3B-Thinking~\cite{yang2025qwen3} under identical HardGen configurations by having each model generate 2,000 trajectories. In \cref{fig:kattempt}, Qwen3-30B-A3B-Thinking achieves the highest correctness rate across all K values, reaching 89\% at K=3. Notably, this model activates only 3B parameters per forward pass through its mixture-of-experts architecture, enabling significantly faster generation than the dense 32B models while delivering superior performance. Its combination of efficient inference, robust multi-step reasoning, and high correctness on function call synthesis makes it the optimal choice for cost-effective large-scale data generation.

\begin{figure}[t]
    \centering
    \includegraphics[width=1.0\linewidth]{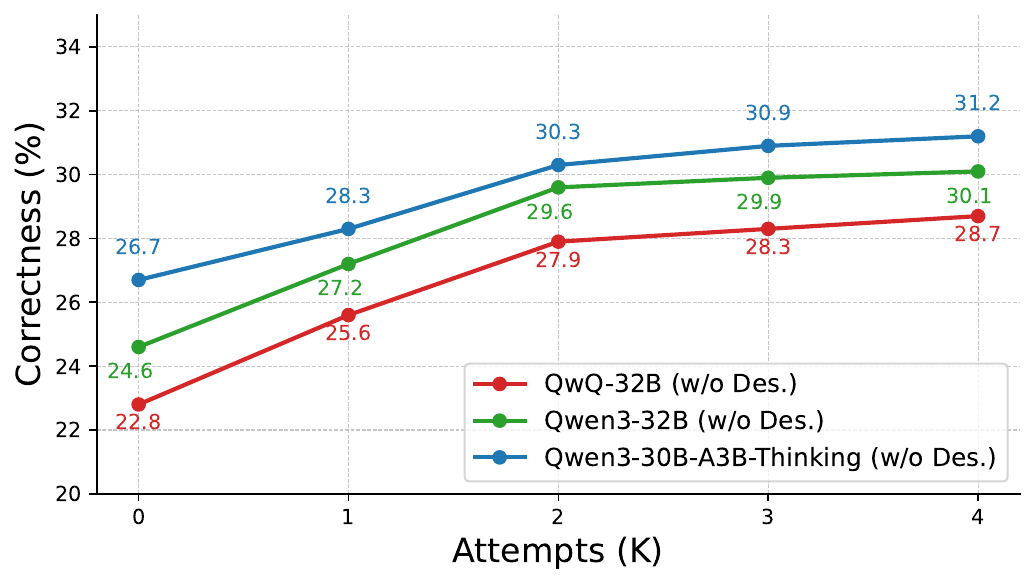}
    \caption{\textbf{Impact of advanced tool descriptions}. Correctness rates of the generator models across attempts ($K$) when advanced tool descriptions are omitted, illustrating the necessity of high-level abstractions for reasoning.}
    \label{fig:des}
\end{figure}

\noindent\textbf{The Necessity of Advanced Tool Description for CoT  .} Compared with the findings in \cref{fig:kattempt},  \cref{fig:des} further provides compelling evidence for the necessity of advanced tool descriptions in CoT generation. Across all models and attempt counts, removing descriptions results in uniformly poor performance, with maximum correctness rates below 32. The limited differentiation between models—only 2.5 percentage points separate the best and worst performers at $K_{\max}=4$, suggests that the correctness of reasoning is low  without adequate tool information. Moreover, the diminishing returns from additional attempts (performance gains $<1$\% after $K=2$) indicate that models cannot iteratively refine their selections through reasoning alone. These findings demonstrate that advanced tool descriptions constitute a fundamental prerequisite rather than a mere performance enhancement for enabling LLMs to engage in rigorous reasoning when confronted with hard queries.

\begin{table}[t]
\centering
\small
\resizebox{0.49\textwidth}{!}{
\begin{tabular}{lccc}
\toprule
\textbf{Agent Backbone} & \textit{Single-Turn}& \textit{Multi-Turn} & \textbf{Overall} \\
\midrule
None (Base Model)& 82.14 & 22.13& 62.13\\
QwQ-32B& 86.19 & 52.37 & 74.92\\
Qwen3-32B & 86.87 & 58.29 & 77.34\\
Qwen3-30B-A3B-Thinking & 87.14 & 63.13 & 79.14 \\
\bottomrule
\end{tabular}}
\caption{\textbf{Impact of data generation backbones}. Performance of the Qwen3-4B model trained on data synthesized by different agent backbones.}
\vspace{-1em}
\label{tab:backbone}
\end{table}

\noindent\textbf{Data Generation with Other Agent Backbones.} To demonstrate the robustness and generalizability of HardGen across different agent backbones, we conduct ablation experiments using alternative model backbones for data generation. As shown in \cref{tab:backbone}, we evaluate three models—Qwen3-32B~\cite{yang2025qwen3}, QwQ-32B~\cite{team2025qwq}, and Qwen3-30B-A3B-Thinking~\cite{yang2025qwen3}, as the agent backbones for synthesizing training data, while keeping all other pipeline components identical. The results demonstrate the strong effectiveness of HardGen for synthetic data generation, yielding substantial improvements over the base model across all agent backbones. These consistent and pronounced gains indicate that HardGen’s strategy of generating challenging, agent-produced synthetic data effectively cultivates complex reasoning capabilities, with performance benefits scaling in tandem with the strength of the generation backbone.

\section{Evaluations}

Recent efforts in benchmarking LLM tool-use have centered on three key axes: scalability, robustness, and realism.

\noindent\textbf{BFCL.} For scalability, the Berkeley Function Calling Leaderboard (BFCL)~\cite{patilberkeley} introduces a novel validation strategy using Abstract Syntax Tree sub-string matching. This approach serves as a proxy for function execution, enabling large-scale, deterministic evaluation across diverse categories, including Single-Turn, Multi-Turn (BFCLv3), and Agentic scenarios ((BFCLv4).

\noindent\textbf{ACEBench.} For robustness, the ACEBench~\cite{chen2025acebench} uses a sandbox environment  that measures models performance on dynamic, simulated tasks. It uses two distinct metrics, End-to-End Accuracy, which compares the final instance attributes of the environment with the target state; and Process Accuracy, which measures the consistency between the actual function call process and the ideal process . This approach is designed to capture task completion in realistic, interactive scenarios.

\noindent\textbf{APIBank.} For realism, the API-Bank~\cite{li2023api} establishes a framework for runnable evaluation, grading tasks into complex, multi-step sequences. The system verifies whether the same database queries or modifications are performed  to ensure the ground-truth outcome is achieved.

\section{Performance on the Held-out Benchmark BFCLv4}
\label{appendix:bfclv4}

\begin{table}[t]
\centering
\small
\resizebox{0.48\textwidth}{!}{
\begin{tabular}{lccc}
\toprule
\textbf{Model} & \textit{Web Search} & \textit{Memory} & \textbf{Overall}\\
\midrule
 Llama-3.1-8B-Instruct & 3.00	&10.75&6.88\\
\midrule
CoALM-8B	&0.00	&2.80&1.40\\
ToolACE-2-8B &8.50	&18.49&13.50	\\
BitAgent-8B	 &4.00	&12.47	&8.24\\
xLAM-2-8b-fc-r 	&6.50	&13.98&10.24\\
\midrule

\rowcolor[HTML]{EAE8FD}+ \textbf{HardGen-RL} &\textbf{16.00}&\textbf{24.84}&\textbf{20.42}\\
\rowcolor[HTML]{EAE8FD}{$\Delta$} & \textcolor{red}{+13.00}         & \textcolor{red}{+14.09}      & \textcolor{red}{+13.54}\\
\bottomrule
\end{tabular}}
\caption{\textbf{Out-of-distribution evaluation on BFCLv4}. Performance of HardGen-RL on two specific agentic tasks, in comparison with other methods built on the same base model.}
\label{tab:bfclv4}
\end{table}


\noindent\textbf{Generalization of Agentic Capabilities.} To assess the transferability of the tool-use skills induced by HardGen, we further evaluate our method on the out-of-distribution BFCLv4 benchmark, comprising two representative agentic tasks: the Search and Memory subsets. From the results shown in \cref{tab:bfclv4}, the HardGen-8B-RL model exhibits remarkable robustness, delivering a substantial uplift over the base model. Specifically, the overall performance surges by \textbf{+13.54} points (rising from 6.88 to 20.42). This advantage is most pronounced in the Memory subset, where accuracy improvement is \textbf{+14.09} (from 10.75 to 24.84), accompanied by a performance boost of \textbf{+13.00} in the Web Search subset (3.00 to 16.00).Crucially, HardGen-RL consistently outperforms all other baselines built on the same backbone. These results strongly suggest that our data generation paradigm effectively instills robust agentic behaviors, providing a solid foundation for future research in agentic reinforcement learning.

\section{Training details}
\label{app:trainingdetails}

\begin{table}[h!]
\centering
\resizebox{0.25\textwidth}{!}{
\begin{tabular}{lc}
\toprule
\textbf{Hyperparameter} & \textbf{Value} \\
\midrule
Batch Size & 1024 \\
Learning Rate & $4\mathrm{e}{-5}$ \\
Max Length & 20480 \\
Epoch Number& 5\\
\bottomrule
\end{tabular}}
\caption{\textbf{Hyperparameters for SFT}. }
\label{tab:SFTsetting}
\end{table}

\noindent\textbf{Supervised Finetuning.} We conduct our supervised finetuning experiments using the open-source Llama Factory library~\cite{zheng2024llamafactory}. The main hyperparameter settings are listed in \cref{tab:SFTsetting}.

\begin{table}[t]
\centering
\resizebox{0.25\textwidth}{!}{
\begin{tabular}{lc}
\toprule
\textbf{Hyperparameter} & \textbf{Value} \\
\midrule
Batch Size & 512 \\
Learning Rate & $1\mathrm{e}{-6}$ \\
KL Coefficient & $1\mathrm{e}{-3}$ \\
Entropy Coefficient & 0 \\
Max Length & 20480 \\
Temperature & 0.7 \\
Epoch Number& 5\\
Rollout Number &16\\
\bottomrule
\end{tabular}}
\caption{\textbf{Hyperparameters for RL training}. }
\label{tab:RLsetting}
\end{table}

\noindent\textbf{RL Training.} We conduct our reinforcement learning (RL) experiments using the open-source Verl library~\cite{sheng2024hybridflow}. To ensure stable and efficient training, we adopt the training settings from in~\cite{zhang2025nemotron}. The key hyperparameter settings are summarized in \cref{tab:RLsetting}.

\noindent\textbf{Reward Design.} We adopt a simple yet effective reward that is widely used in prior works~\cite{zhang2025nemotron,yu2025dapo,zeng2025glm}. Given a query $q$ with reference answer $g$, the model's output $o$ is evaluated as follows. If $o$ contains a tool call, it is considered correct only when it can be successfully parsed into valid function calls with proper parameters, exactly matches $g$, and follows the prescribed reasoning template. In contrast, if $g$ does not contain a tool call, then $o$ is considered correct only when it contains no valid function calls (\textit{i.e.}, is free-form text) while still adhering to the required reasoning template. The reward is thus defined as:
\begin{equation}
\text{Reward} =
\begin{cases}
1, & \text{format correct \& answer correct,}  \\
0, & \text{otherwise.}
\end{cases}
\end{equation}

This binary reward emphasizes the holistic integrity of the output, enforcing not only semantic correctness but also strict structural compliance, which is essential for reliable downstream execution. In non-tool-calling cases, the absence of valid function calls implicitly verifies that the model produces a purely textual response and prevents spurious or unnecessary tool invocations.

\section{Manual Annotations}
\label{appendix:manualann}

\noindent\textbf{Impact of the Advanced Tools.} Our proposed method introduces advanced tool ($T_{adv}$) to help LLMs bridge logical jumps during hard query synthesis. We present the proportion of challenging queries across all model scales in \cref{tab:advtool}. However, the \emph{difficulty} judgments are the product of automatic annotation (GPT-4o), and this evaluation task moves beyond simple surface-level text generation judgment. To further evaluate whether the model successfully navigate the hard logical jump that the non-advanced tool model failed, three PhD students in NLP field (three of the authors) form an annotation team to annotate the samples anew using logical jumps as the indicator. If there are differences between two annotations on a sample, the third annotation will be introduced to determine the final decision. 
From \cref{tab:advtoolhuman}, we can observe a strong consistency between the automatic and manual annotations, with no significant differences between the human and model conclusions. Overall, these results demonstrate that the introduction of $T_{adv}$ effectively enhances the model’s ability to construct logically jumping queries, with consistent benefits observed across model scales.


\begin{table}[t]
\centering
\resizebox{\linewidth}{!}{
    \begin{tabular}{lcccccc}
    \toprule
    \multirow{2}{*}{\textbf{Models}} & \multicolumn{2}{c}{\textit{Consistency}} & \multicolumn{2}{c}{\textit{Pearson Corr.}} & \multicolumn{2}{c}{\textbf{Human Preference}} \\
    \cmidrule(lr){2-3} \cmidrule(lr){4-5} \cmidrule(lr){6-7}
       & \textit{w/o $T_{adv}$} & \textit{w/ $T_{adv}$} & \textit{w/o $T_{adv}$} & \textit{w/ $T_{adv}$} & \textit{w/o $T_{adv}$} &\textit{w/ $T_{adv}$}\\
    \midrule
    \makecell[l]{Qwen3-30B-\\A3B-Thinking} &0.86  & 0.87 &0.72  & 0.74 &182  & \textbf{218} \\
    \midrule
    \makecell[l]{Qwen3-235B-\\A22B-Thinking} &  0.89 &0.91 &0.78  &0.82  &163 &  \textbf{237} \\
    \midrule
    \makecell[l]{DeepSeek-\\V3.1 (671B)} &  0.92 & 0.95& 0.84  &0.90 &129  & \textbf{271} \\
    \bottomrule
    \end{tabular}
}
\caption{\textbf{Agreement between manual and automatic annotations.} We report the Consistency rates and Pearson correlations between the two annotation methods, alongside the Human Preference.}
\label{tab:advtoolhuman}
\end{table}

\section{Results on Other Models}
\label{appendix:othermodels}

\begin{table}[t]
\centering
\small
\begin{tabular}{lccc}
\toprule
\textbf{Models} & \textit{Single-Turn} & \textit{Multi-Turn} & \textbf{Overall}\\
\midrule
Llama-3.1-8B-Inst & 77.38& 11.12& 55.29\\
\textbf{+HardGen-RL}  &84.55&53.10&74.07\\
\midrule
Llama-3.2-3B-Inst & 70.50 & 4.00& 48.33\\
\textbf{+HardGen-RL} &84.05&40.13&67.57\\
\bottomrule
\end{tabular}
\caption{\textbf{Generalization to the Llama model family}. Performance comparison of Llama-3.1-8B and Llama-3.2-3B models before and after HardGen-RL training on BFCLv3 Single-Turn and Multi-Turn tasks.}
\vspace{-1em}
\label{tab:llamaModels}
\end{table}

\noindent\textbf{Results on Llama3 Models.} To assess the generalizability of our approach beyond Qwen models, we apply HardGen-RL to two Llama-3 variants, Llama-3.1-8B-Instruct and Llama-3.2-3B-Instruct. As shown in \cref{tab:llamaModels}, both models demonstrate substantial improvements after reinforcement learning. Llama-3.1-8B-Instruct achieves an overall accuracy of 74.07 (\textbf{+18.78}), with particularly strong gains on multi-turn interactions (53.10, \textbf{+41.98}). Similarly, Llama-3.2-3B-Instruct improves from 48.33 to 67.57 overall (\textbf{+19.24}), with multi-turn performance increasing from 4.00 to 40.13 (\textbf{+36.13}).  These results confirm that our approach effectively transfers across model families and parameter scales, establishing HardGen as a robust data generation framework for enhancing tool-use capabilities beyond the Qwen architecture.

\section{Construction of the API Graph}
\label{app:graph}

\paragraph{Structure of API Graph.}
The API Graph $\mathcal{G} = (\mathcal{T}, \mathcal{D}, \mathcal{P})$ encodes three types of information critical for hard trace generation: (1) \emph{Failure Tool Set} $\mathcal{T}$: the set of 1,204 failure APIs identified through self-evaluation; (2) \emph{Dependencies} $\mathcal{D}$: directed edges representing prerequisite relationships, where $(T_i, T_j) \in \mathcal{D}$ indicates that $T_j$ requires $T_i$ to be executed first; and (3) \emph{Parameter Constraints} $\mathcal{P}$: specifications defining valid parameter ranges, types, and inter-tool parameter mappings (e.g., output type of $T_i$ must match input type of $T_j$). Parameter constraints $\mathcal{P}$ are extracted from tool API schemas (type signatures, value ranges, required fields).

\paragraph{Update of API Graph.}
The API graph is updated  through execution feedback from the environment. After each tool call, we analyze the feedback to identify dependencies. When tool $T_j$ serves as a preceding tool of of $T_i$, the execution of $T_j$ activates $T_i$, making it eligible for selection in subsequent steps. Concurrently, parameter constraints in $\mathcal{P}$ are refined by tracking value validity ranges and type requirements observed during execution. This enables the graph to capture implicit value-level dependencies that extend beyond simple tool-set inclusion relationships. Through this continual update process, the API graph progressively refines both structural and parameter-level dependencies, thereby biasing subsequent sampling toward tool sequences that are both executable and challenging.

\section{Supplementary Case}
\label{app:additional_results}

This section presents a case study of HardGen, detailing the synthesis process along with a complete trajectory.

\noindent\textbf{Hard Query Construction with Logical Jump.} \cref{fig:hard_query_construction} illustrates the construction of hard queries with logical jumps. Unlike the previous methods, which explicitly instruct the model to first check zip codes before purchasing tickets, the hard query directly requests ticket purchase between city names without specifying intermediate steps. This formulation requires the model to autonomously infer the necessary tool chain—recognizing that city names must first be converted to zip codes via \texttt{get\_zipcode} before invoking \texttt{buy\_tickets}. The advanced tool \texttt{buy\_tickets\_adv} demonstrates the desired end-to-end capability of purchasing tickets directly from city names, representing the ideal abstraction, a single function that internally handles the multi-step process.

\noindent\textbf{Model Reasoning for Hard Query.} \cref{fig:model_reasoning} illustrates the model's reasoning process for a hard query. Given the task to purchase tickets between two cities by name, the model correctly recognizes the mismatch between the query (city names) and available tool inputs (zip codes). In its internal reasoning, the model analyzes the available tools, identifies the dependency structure, and decomposes the task into three steps: (1) retrieve the zip code for Rivermist, (2) retrieve the zip code for Stonebrook, and (3) purchase tickets using both zip codes. The model then executes this planned sequence through appropriate tool calls, successfully completing the multi-step task without explicit instructions.

\noindent\textbf{Complete Trajectory.} \cref{fig:trajectory_turn1,fig:trajectory_turn2} illustrate a complete multi-turn trajectory demonstrating sequential reasoning and context retention. In Turn 1, the model is asked to determine the working directory and search for all files recursively. The model correctly selects \texttt{pwd} and \texttt{find} tools, retrieves the directory structure, and summarizes the results. In Turn 2, building on the previous context, the user requests file contents from subdirectories identified in Turn 1. The model recognizes tool constraints—that \texttt{cat} and \texttt{tail} only operate within the current directory—and constructs a four-step plan involving directory navigation (\texttt{cd}) to access the required files. This example demonstrates the model's ability to maintain context across turns and adapt its strategy based on tool limitations.

\definecolor{thinkcolor}{RGB}{138, 43, 226}
\definecolor{toolcolor}{RGB}{34, 139, 34}
\definecolor{toolresponsecolor}{RGB}{70, 130, 180}
\definecolor{answercolor}{RGB}{25, 25, 112}
\definecolor{usercolor}{RGB}{105, 105, 105}
\definecolor{groundtruthcolor}{RGB}{139, 69, 19}
\definecolor{querycolor}{RGB}{70, 130, 180}
\definecolor{tracecolor}{RGB}{34, 139, 34}

\newcommand{\thinking}[1]{\textcolor{thinkcolor}{\textbf{#1}}}
\newcommand{\toolcall}[1]{\textcolor{toolcolor}{\textbf{#1}}}
\newcommand{\toolresponse}[1]{\textcolor{toolresponsecolor}{\textbf{#1}}}
\newcommand{\finalanswer}[1]{\textcolor{answercolor}{\textbf{#1}}}

\begin{figure*}[b]
\centering
\begin{tcolorbox}[
    colback=white,
    colframe=toolresponsecolor!80,
    boxrule=1.5pt,
    arc=4pt,
    left=8pt,
    right=8pt,
    top=6pt,
    bottom=6pt,
    width=0.96\textwidth,
    title={\textbf{Case: Hard Query Construction with Logical Jump}},
    fonttitle=\small\bfseries,
    coltitle=white,
    colbacktitle=toolresponsecolor!80
]
\small

\begin{tcolorbox}[
    colback=tracecolor!8,
    colframe=tracecolor!60,
    boxrule=1pt,
    arc=3pt,
    left=6pt,
    right=6pt,
    top=4pt,
    bottom=4pt
]
\textbf{Original Trace:}\\[0.3em]
\texttt{get\_zipcode(city="Rivermist")} $\rightarrow$ \texttt{"83214"} \\
\texttt{get\_zipcode(city="Stonebrook")} $\rightarrow$ \texttt{"74532"} \\
\texttt{buy\_tickets(zipcodeA="83214", zipcodeB="74532")} $\rightarrow$ \texttt{"ticket\_id": "14589"}
\end{tcolorbox}

\begin{tcolorbox}[
    colback=groundtruthcolor!8,
    colframe=groundtruthcolor!60,
    boxrule=1pt,
    arc=3pt,
    left=6pt,
    right=6pt,
    top=4pt,
    bottom=4pt
]
\textbf{Advanced Tool:} \texttt{buy\_tickets\_adv(cityA, cityB) -> ticket\_id}\\
\textbf{Description:} Purchase air tickets between two cities by city names, returning the purchased ticket information.
\end{tcolorbox}

\begin{tcolorbox}[
    colback=querycolor!8,
    colframe=querycolor!60,
    boxrule=1pt,
    arc=3pt,
    left=6pt,
    right=6pt,
    top=4pt,
    bottom=4pt
]
\textbf{Easy Query:}\\
Please check the zip code of the city ``Rivermist'' and ``Stonebrook'' first, then purchase air tickets between the two cities according to the zip codes you checked.

\textbf{Hard Query:}\\
Please purchase air tickets between the city ``Rivermist'' and the city ``Stonebrook''.
\end{tcolorbox}

\begin{tcolorbox}[
    colback=thinkcolor!8,
    colframe=thinkcolor!60,
    boxrule=1pt,
    arc=3pt,
    left=6pt,
    right=6pt,
    top=4pt,
    bottom=4pt
]
\textbf{Hint:}\\
Purchase air tickets between two cities by city names, returning the purchased ticket information.
\end{tcolorbox}

\end{tcolorbox}
\caption{\textbf{Case Study: Hard Query Construction}. Illustration of the data synthesis process where an original execution trace is abstracted into an ``Advanced Tool'' to generate a Hard Query that omits explicit intermediate steps (zipcode lookup).}
\label{fig:hard_query_construction}
\end{figure*}

\begin{figure*}[t]
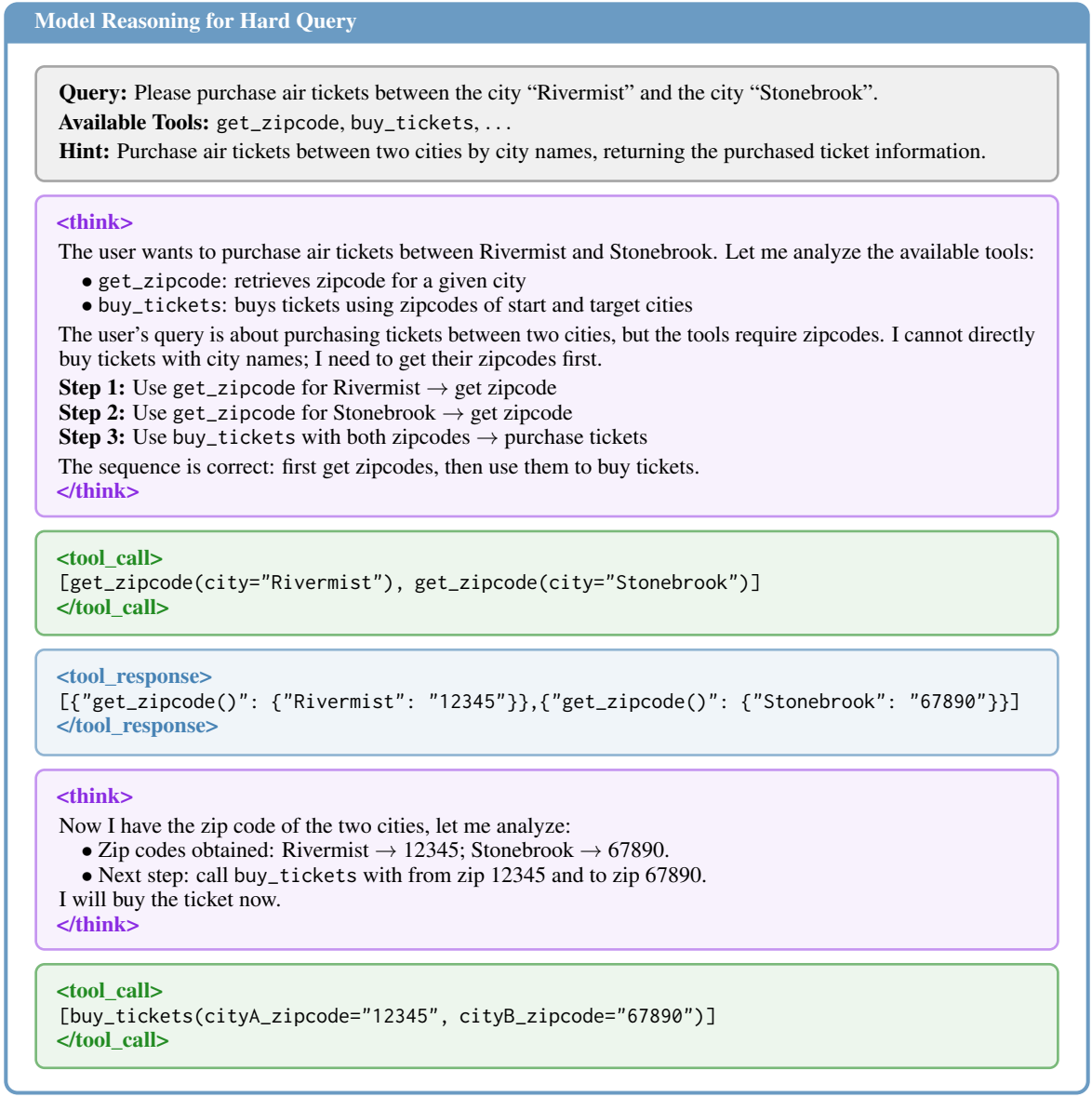

\centering
\begin{tcolorbox}[
    colback=white,
    colframe=toolresponsecolor!80,
    boxrule=1.5pt,
    arc=4pt,
    left=8pt,
    right=8pt,
    top=6pt,
    bottom=6pt,
    width=0.96\textwidth,
    title={\textbf{Model Reasoning for Hard Query}},
    fonttitle=\small\bfseries,
    coltitle=white,
    colbacktitle=toolresponsecolor!80
]
\small

\begin{tcolorbox}[
    colback=usercolor!10,
    colframe=usercolor!60,
    boxrule=1pt,
    arc=3pt,
    left=6pt,
    right=6pt,
    top=4pt,
    bottom=4pt
]
\textbf{Query:} Please purchase air tickets between the city ``Rivermist'' and the city ``Stonebrook''.\\[0.2em]
\textbf{Available Tools:} \texttt{get\_zipcode}, \texttt{buy\_tickets}, $\ldots$\\[0.2em]
\textbf{Hint:} Purchase air tickets between two cities by city names, returning the purchased ticket information.
\end{tcolorbox}
\begin{tcolorbox}[
    colback=thinkcolor!6,
    colframe=thinkcolor!50,
    boxrule=1pt,
    arc=3pt,
    left=6pt,
    right=6pt,
    top=4pt,
    bottom=4pt
]
\thinking{<think>}\\[0.2em]
The user wants to purchase air tickets between Rivermist and Stonebrook. Let me analyze the available tools:\\[0.2em]
\hspace*{1em}$\bullet$ \texttt{get\_zipcode}: retrieves zipcode for a given city\\
\hspace*{1em}$\bullet$ \texttt{buy\_tickets}: buys tickets using zipcodes of start and target cities\\[0.2em]
The user's query is about purchasing tickets between two cities, but the tools require zipcodes. I cannot directly buy tickets with city names; I need to get their zipcodes first.\\[0.2em]
\textbf{Step 1:} Use \texttt{get\_zipcode} for Rivermist $\rightarrow$ get zipcode\\
\textbf{Step 2:} Use \texttt{get\_zipcode} for Stonebrook $\rightarrow$ get zipcode\\
\textbf{Step 3:} Use \texttt{buy\_tickets} with both zipcodes $\rightarrow$ purchase tickets\\[0.2em]
The sequence is correct: first get zipcodes, then use them to buy tickets.\\
\thinking{</think>}
\end{tcolorbox}
\begin{tcolorbox}[
    colback=toolcolor!8,
    colframe=toolcolor!60,
    boxrule=1pt,
    arc=3pt,
    left=6pt,
    right=6pt,
    top=4pt,
    bottom=4pt
]
\toolcall{<tool\_call>}\\
\texttt{[get\_zipcode(city="Rivermist"), get\_zipcode(city="Stonebrook")]}\\
\toolcall{</tool\_call>}
\end{tcolorbox}
\begin{tcolorbox}[
    colback=toolresponsecolor!8,
    colframe=toolresponsecolor!60,
    boxrule=1pt,
    arc=3pt,
    left=6pt,
    right=6pt,
    top=4pt,
    bottom=4pt
]
\toolresponse{<tool\_response>}\\
\texttt{[\{"get\_zipcode()": \{"Rivermist": "12345"\}\},}\texttt{\{"get\_zipcode()": \{"Stonebrook": "67890"\}\}]}\\
\toolresponse{</tool\_response>}
\end{tcolorbox}

\begin{tcolorbox}[
    colback=thinkcolor!6,
    colframe=thinkcolor!50,
    boxrule=1pt,
    arc=3pt,
    left=6pt,
    right=6pt,
    top=4pt,
    bottom=4pt
]
\thinking{<think>}\\[0.2em]
Now I have the zip code of the two cities, let me  analyze:\\
\hspace*{1em}$\bullet$ Zip codes obtained: Rivermist → 12345; Stonebrook → 67890.\\
\hspace*{1em}$\bullet$ Next step: call \texttt{buy\_tickets} with from zip 12345 and to zip 67890.\\
I will buy the ticket now.\\
\thinking{</think>}
\end{tcolorbox}

\begin{tcolorbox}[
    colback=toolcolor!8,
    colframe=toolcolor!60,
    boxrule=1pt,
    arc=3pt,
    left=6pt,
    right=6pt,
    top=4pt,
    bottom=4pt
]
\toolcall{<tool\_call>}\\
\texttt{[buy\_tickets(cityA\_zipcode="12345", cityB\_zipcode="67890")]}\\
\toolcall{</tool\_call>}
\end{tcolorbox}
\end{tcolorbox}
\caption{\textbf{Case Study: Reasoning process for a Hard Query}. Visualization of the model's Chain-of-Thought (CoT) process, showing how it decomposes a high-level query into necessary primitive tool calls (\textit{e.g.}, retrieving zip codes before purchase).}
\label{fig:model_reasoning}
\end{figure*}

\begin{figure*}[t]
\centering
\begin{tcolorbox}[
    colback=white,
    colframe=toolresponsecolor!80,
    boxrule=1.5pt,
    arc=4pt,
    left=8pt,
    right=8pt,
    top=6pt,
    bottom=6pt,
    width=0.96\textwidth,
    title={\textbf{Complete Trajectory -- Turn 1}},
    fonttitle=\small\bfseries,
    coltitle=white,
    colbacktitle=toolresponsecolor!80
]
\small

\begin{tcolorbox}[
    colback=usercolor!10,
    colframe=usercolor!60,
    boxrule=1pt,
    arc=3pt,
    left=6pt,
    right=6pt,
    top=4pt,
    bottom=4pt
]
\textbf{User:} Determine the current working directory path, and then search for all files and directories in the current directory and its subdirectories.
\end{tcolorbox}

\begin{tcolorbox}[
    colback=thinkcolor!6,
    colframe=thinkcolor!50,
    boxrule=1pt,
    arc=3pt,
    left=6pt,
    right=6pt,
    top=4pt,
    bottom=4pt
]
\thinking{<think>}\\[0.2em]
The user wants to: (1) determine current working directory, (2) search all files/directories recursively.\\[0.2em]
\hspace*{1em}$\bullet$ \texttt{pwd}: returns current directory path (no parameters needed)\\
\hspace*{1em}$\bullet$ \texttt{find}: searches files/directories; \texttt{path="."} for current dir, \texttt{name=None} for all items\\[0.2em]
The \texttt{ls} command only lists current directory, not subdirectories, so \texttt{find} is the right choice.\\
\thinking{</think>}
\end{tcolorbox}

\begin{tcolorbox}[
    colback=toolcolor!8,
    colframe=toolcolor!60,
    boxrule=1pt,
    arc=3pt,
    left=6pt,
    right=6pt,
    top=4pt,
    bottom=4pt
]
\toolcall{<tool\_call>}\\
\texttt{[pwd(), find(path=".")]}\\
\toolcall{</tool\_call>}
\end{tcolorbox}

\begin{tcolorbox}[
    colback=toolresponsecolor!8,
    colframe=toolresponsecolor!60,
    boxrule=1pt,
    arc=3pt,
    left=6pt,
    right=6pt,
    top=4pt,
    bottom=4pt
]
\toolresponse{<tool\_response>}\\
\texttt{[\{'pwd()': '\{"current\_working\_directory": "/workspace"\}'\},}\\
\texttt{\ \{'find(path=".")': '\{"matches": ["./processed", "./data", "./data/raw",}\\
\texttt{\ "./data/raw/dtpsp.json", "./data/unggv.json", ...]\}'\}]}\\
\toolresponse{</tool\_response>}
\end{tcolorbox}

\begin{tcolorbox}[
    colback=answercolor!8,
    colframe=answercolor!60,
    boxrule=1pt,
    arc=3pt,
    left=6pt,
    right=6pt,
    top=4pt,
    bottom=4pt
]
\textbf{Assistant Summary:} The current working directory is \texttt{/workspace}. The search found the following items: \texttt{["./processed", "./data", "./data/raw", "./data/raw/dtpsp.json", ...]}
\end{tcolorbox}

\end{tcolorbox}
\caption{\textbf{Qualitative Example: Multi-turn Directory Exploration (Turn 1)}. A trajectory step where the model correctly selects the \texttt{pwd} and \texttt{find} commands to satisfy a recursive search requirement.}
\label{fig:trajectory_turn1}
\end{figure*}

\begin{figure*}[t]
\centering
\begin{tcolorbox}[
    colback=white,
    colframe=toolresponsecolor!80,
    boxrule=1.5pt,
    arc=4pt,
    left=8pt,
    right=8pt,
    top=6pt,
    bottom=6pt,
    width=0.96\textwidth,
    title={\textbf{Complete Trajectory -- Turn 2}},
    fonttitle=\small\bfseries,
    coltitle=white,
    colbacktitle=toolresponsecolor!80
]
\small

\begin{tcolorbox}[
    colback=usercolor!10,
    colframe=usercolor!60,
    boxrule=1pt,
    arc=3pt,
    left=6pt,
    right=6pt,
    top=4pt,
    bottom=4pt
]
\textbf{User:} Display the contents of the file named \texttt{unggv.json} in the data directory, and then display the last 7 lines of the file named \texttt{dtpsp.json} in the raw subdirectory.
\end{tcolorbox}

\begin{tcolorbox}[
    colback=thinkcolor!6,
    colframe=thinkcolor!50,
    boxrule=1pt,
    arc=3pt,
    left=6pt,
    right=6pt,
    top=4pt,
    bottom=4pt
]
\thinking{<think>}\\[0.2em]
Current directory: \texttt{/workspace}. Need to access files in subdirectories.\\
\textbf{Constraints:} \texttt{cat} and \texttt{tail} only work on files in the current directory.\\[0.2em]
\textbf{Plan:}\\
\hspace*{1em}(1) \texttt{cd data} $\rightarrow$ change to data directory\\
\hspace*{1em}(2) \texttt{cat unggv.json} $\rightarrow$ display file contents\\
\hspace*{1em}(3) \texttt{cd raw} $\rightarrow$ change to raw subdirectory\\
\hspace*{1em}(4) \texttt{tail dtpsp.json lines=7} $\rightarrow$ display last 7 lines\\
\thinking{</think>}
\end{tcolorbox}

\begin{tcolorbox}[
    colback=toolcolor!8,
    colframe=toolcolor!60,
    boxrule=1pt,
    arc=3pt,
    left=6pt,
    right=6pt,
    top=4pt,
    bottom=4pt
]
\toolcall{<tool\_call>}\\
\texttt{[cd(folder="data"), cat(file\_name="unggv.json"), cd(folder="raw"),}\\
\texttt{\ tail(file\_name="dtpsp.json", lines=7)]}\\
\toolcall{</tool\_call>}
\end{tcolorbox}

\begin{tcolorbox}[
    colback=toolresponsecolor!8,
    colframe=toolresponsecolor!60,
    boxrule=1pt,
    arc=3pt,
    left=6pt,
    right=6pt,
    top=4pt,
    bottom=4pt
]
\toolresponse{<tool\_response>}\\
\texttt{[\{'cd(folder="data")': '\{"current\_working\_directory": "data"\}'\},}\\
\texttt{\ \{'cat(file\_name="unggv.json")': '\{"file\_content": "\{"data": \{"id": "94"\}\}"\}'\},}\\
\texttt{\ \{'cd(folder="raw")': '\{"current\_working\_directory": "raw"\}'\},}\\
\texttt{\ \{'tail(...)': '\{"last\_lines": "\{"data": \{"id": "27"\}\}"\}'\}]}\\
\toolresponse{</tool\_response>}
\end{tcolorbox}

\begin{tcolorbox}[
    colback=answercolor!8,
    colframe=answercolor!60,
    boxrule=1pt,
    arc=3pt,
    left=6pt,
    right=6pt,
    top=4pt,
    bottom=4pt
]
\textbf{Assistant Summary:}\\
The contents of ``unggv.json'': \texttt{\{"data": \{"id": "94", "type": "product"\}\}}\\
The last 7 lines of ``dtpsp.json'': \texttt{\{"data": \{"id": "27", "type": "user"\}\}}
\end{tcolorbox}

\end{tcolorbox}
\caption{\textbf{Qualitative Example: Context-Dependent File Retrieval (Turn 2)}. A subsequent trajectory step demonstrating context retention, where the model navigates directories (\texttt{cd}) to access files based on tool constraints.}
\label{fig:trajectory_turn2}
\end{figure*}

\section{Prompts}
\label{appendix:prompt}
This section provides the prompts used in the HardGen pipeline. Our framework employs four specialized agents, each with carefully designed prompts to fulfill specific roles in the data generation process. The Tool Maker (\cref{fig:tool_maker_prompt}) synthesizes advanced tools from multi-step execution hard traces by abstracting primitive tool sequences into high-level operations. The Hard Query Generator (\cref{fig:hard_query_prompt}) creates challenging queries that require implicit logical bridging, forcing models to infer necessary intermediate steps rather than following explicit instructions. During the reasoning refinement phase, the Reasoner (\cref{fig:reasoning_agent_prompt} and \cref{fig:refinement_agent_prompt}) attempts to solve hard queries with hints from advanced tool descriptions, while the Verifier (\cref{fig:verifier_agent_prompt}) analyzes incorrect attempts and provides targeted corrective feedback without revealing answers directly. These prompts collectively enable HardGen's closed-loop refinement mechanism, where each component builds upon the outputs of previous stages. The Tool Maker and Hard Query Generator transform executable traces into challenging learning scenarios, while the Reasoner-Verifier loop iteratively refines chain-of-thought reasoning toward correct solutions. All prompts are designed to be model-agnostic and can be adapted to different LLM architectures. The structured output format ensures consistency and inherent executability, enabling direct execution without additional validation.

\begin{figure*}[t]
\centering
\begin{tcolorbox}[
    colback=white,
    colframe=purple!80,
    boxrule=1.5pt,
    arc=4pt,
    left=8pt,
    right=8pt,
    top=6pt,
    bottom=6pt,
    width=0.96\textwidth,
    title={\textbf{Tool Maker Agent Prompt}},
    fonttitle=\small\bfseries,
    coltitle=white,
    colbacktitle=purple!80
]
\small

\begin{tcolorbox}[
    colback=gray!10,
    colframe=gray!60,
    boxrule=1pt,
    arc=3pt,
    left=6pt,
    right=6pt,
    top=4pt,
    bottom=4pt
]
\textbf{System Prompt:}\\[0.3em]
You are a Tool Maker agent responsible for synthesizing advanced tools from execution traces.

\vspace{0.3em}
\textbf{Given the following execution trace:}\\
\texttt{\{execution\_trace\}}

\vspace{0.3em}
\textbf{Trace Details:}
\begin{itemize}[leftmargin=1.5em, itemsep=0.1em]
    \item Tool Call 1: \texttt{\{tool\_1\}(\{params\_1\})} $\rightarrow$ \texttt{\{result\_1\}}
    \item Tool Call 2: \texttt{\{tool\_2\}(\{params\_2\})} $\rightarrow$ \texttt{\{result\_2\}}
    \item \ldots
    \item Tool Call M: \texttt{\{tool\_M\}(\{params\_M\})} $\rightarrow$ \texttt{\{result\_M\}}
\end{itemize}
\end{tcolorbox}

\vspace{0.3em}

\begin{tcolorbox}[
    colback=orange!8,
    colframe=orange!60,
    boxrule=1pt,
    arc=3pt,
    left=6pt,
    right=6pt,
    top=4pt,
    bottom=4pt
]
\textbf{Your task:}\\[0.2em]
Create an advanced tool that abstracts this multi-step execution sequence into a single high-level operation.

\vspace{0.3em}
\textbf{Requirements:}
\begin{enumerate}[leftmargin=1.5em, itemsep=0.2em]
    \item The advanced tool should encapsulate ALL steps in the trace
    \item It should have a clear, intuitive name describing the end-to-end functionality
    \item Parameters should be high-level inputs (not intermediate values)
    \item The description should explain the overall goal, not individual steps
\end{enumerate}

\vspace{0.3em}
\textbf{Output format:}
\begin{verbatim}
{
  "advanced_tool_name": "descriptive_name",
  "parameters": 
  [
    {"name": "param1", "type": "type", 
     "description": "what it represents"
    }
  ]
}
\end{verbatim}
\end{tcolorbox}

\vspace{0.3em}

\begin{tcolorbox}[
    colback=red!8,
    colframe=red!60,
    boxrule=1pt,
    arc=3pt,
    left=6pt,
    right=6pt,
    top=4pt,
    bottom=4pt
]
\textbf{Example:}

\vspace{0.2em}
\colorbox{blue!15}{\textbf{Original trace:}}\\
\texttt{get\_zipcode(city="A")} $\rightarrow$ \texttt{zipA}\\
\texttt{get\_zipcode(city="B")} $\rightarrow$ \texttt{zipB}\\
\texttt{buy\_tickets(zipA, zipB)} $\rightarrow$ \texttt{ticket\_id}

\vspace{0.3em}
\colorbox{green!20}{\textbf{Advanced tool created:}}\\
\texttt{buy\_tickets\_adv(cityA, cityB)} $\rightarrow$ \texttt{ticket\_id}

\vspace{0.3em}
\textbf{Description:} \textit{``Purchase air tickets between two cities by city names, returning the purchased ticket information.''}

\vspace{0.2em}
\textit{The advanced tool hides the intermediate step of zipcode lookup, allowing users to work directly with high-level city names.}
\end{tcolorbox}

\end{tcolorbox}
\caption{\textbf{Prompt template for the Tool Maker Agent}. Instructions used to synthesize advanced tools from multi-step execution traces.}
\label{fig:tool_maker_prompt}
\end{figure*}

\begin{figure*}[t]
\centering
\begin{tcolorbox}[
    colback=white,
    colframe=purple!80,
    boxrule=1.5pt,
    arc=4pt,
    left=8pt,
    right=8pt,
    top=6pt,
    bottom=6pt,
    width=0.96\textwidth,
    title={\textbf{Hard Query Generator Prompt}},
    fonttitle=\small\bfseries,
    coltitle=white,
    colbacktitle=purple!80
]
\small

\begin{tcolorbox}[
    colback=gray!10,
    colframe=gray!60,
    boxrule=1pt,
    arc=3pt,
    left=6pt,
    right=6pt,
    top=4pt,
    bottom=4pt
]
\textbf{System Prompt:}\\[0.3em]
You are a Hard Query Generator agent.

\vspace{0.3em}
\textbf{Given the following advanced tool:}\\
\texttt{\{advanced\_tool\_specification\}}

\vspace{0.3em}
\textbf{Advanced Tool:} \texttt{\{tool\_name\}}\\
\textbf{Parameters:} \texttt{\{parameters\}}\\
\textbf{Description:} \textit{\{description\}}\\
\end{tcolorbox}

\vspace{0.3em}

\begin{tcolorbox}[
    colback=orange!8,
    colframe=orange!60,
    boxrule=1pt,
    arc=3pt,
    left=6pt,
    right=6pt,
    top=4pt,
    bottom=4pt
]
\textbf{Your task:}\\[0.2em]
Generate a challenging query that:
\begin{enumerate}[leftmargin=1.5em, itemsep=0.2em]
    \item Requires the use of this advanced tool's functionality
    \item Does NOT explicitly mention the intermediate steps
    \item Forces the model to perform implicit logical bridging
    \item Uses high-level language that matches the advanced tool's abstraction level
\end{enumerate}

\vspace{0.3em}
\textbf{Requirements for the hard query:}
\begin{itemize}[leftmargin=1.5em, itemsep=0.2em]
    \item[\ding{55}] Do NOT say ``first do X, then do Y, then do Z''
    \item[\ding{55}] Do NOT mention the primitive tools by name
    \item[\ding{51}] DO frame the request at the level of the end goal
    \item[\ding{51}] The query should appear simple but require complex multi-step reasoning
\end{itemize}
\end{tcolorbox}

\vspace{0.3em}

\begin{tcolorbox}[
    colback=red!8,
    colframe=red!60,
    boxrule=1pt,
    arc=3pt,
    left=6pt,
    right=6pt,
    top=4pt,
    bottom=4pt
]
\textbf{Example Comparison:}

\vspace{0.2em}
\colorbox{red!20}{\textbf{BAD (easy) query:}}\\
``Please check the zip code of Rivermist and Stonebrook first, then purchase air tickets between the two cities according to the zip codes you checked.''

\vspace{0.3em}
\colorbox{green!20}{\textbf{GOOD (hard) query:}}\\
``Please purchase air tickets between the city Rivermist and the city Stonebrook.''

\vspace{0.3em}
\textit{The GOOD query requires the model to infer: need zipcodes $\rightarrow$ must call get\_zipcode twice $\rightarrow$ then call buy\_tickets}
\end{tcolorbox}

\end{tcolorbox}
\caption{\textbf{Prompt template for the Hard Query Generator}. Instructions used to create challenging queries that necessitate implicit logical bridging based on advanced tool specifications.}
\label{fig:hard_query_prompt}
\end{figure*}

\begin{figure*}[t]
\centering
\begin{tcolorbox}[
    colback=white,
    colframe=purple!80,
    boxrule=1.5pt,
    arc=4pt,
    left=8pt,
    right=8pt,
    top=6pt,
    bottom=6pt,
    width=0.96\textwidth,
    title={\textbf{Reasoning Agent Prompt (Initial Attempt)}},
    fonttitle=\small\bfseries,
    coltitle=white,
    colbacktitle=purple!80
]
\small

\begin{tcolorbox}[
    colback=gray!10,
    colframe=gray!60,
    boxrule=1pt,
    arc=3pt,
    left=6pt,
    right=6pt,
    top=4pt,
    bottom=4pt
]
\textbf{System Prompt:}\\[0.3em]
You are a reasoning agent tasked with solving tool-use queries.

\vspace{0.3em}
\textbf{Query:} \texttt{\{hard\_query\}}

\vspace{0.3em}
\textbf{Available Tools:}\\
\texttt{\{tool\_descriptions\}}

\vspace{0.3em}
\textbf{Hint (Advanced Tool Capability):}\\
\texttt{\{advanced\_tool\_description\}}

\vspace{0.2em}
\textit{Note: The hint describes a high-level capability. You must use the available primitive tools to achieve this capability.}
\end{tcolorbox}

\vspace{0.3em}

\begin{tcolorbox}[
    colback=orange!8,
    colframe=orange!60,
    boxrule=1pt,
    arc=3pt,
    left=6pt,
    right=6pt,
    top=4pt,
    bottom=4pt
]
\textbf{Your task:}
\begin{enumerate}[leftmargin=1.5em, itemsep=0.2em]
    \item Analyze the query and understand what needs to be accomplished
    \item Identify which tools are needed and in what order
    \item Plan the execution sequence considering tool dependencies
    \item Generate appropriate function calls with correct parameters
\end{enumerate}

\vspace{0.3em}

\textbf{Output format:}

\footnotesize\begin{verbatim}
<think>
[Your reasoning process]
</think>

<tool_call>
[generated_function_call_1, 
 generated_function_call_2, ...]
</tool_call>
\end{verbatim}
\end{tcolorbox}

\vspace{0.3em}

\begin{tcolorbox}[
    colback=red!8,
    colframe=red!60,
    boxrule=1pt,
    arc=3pt,
    left=6pt,
    right=6pt,
    top=4pt,
    bottom=4pt
]
\textbf{Remember:}
\begin{itemize}[leftmargin=1.5em, itemsep=0.2em]
    \item Think carefully about implicit dependencies
    \item Check if tool prerequisites are satisfied
    \item Ensure parameters are correctly mapped between tools
\end{itemize}
\end{tcolorbox}

\end{tcolorbox}
\caption{\textbf{Prompt template for the Reasoner Agent (Initial Attempt)}. Instructions for solving tool-use queries using hints from advanced tool descriptions.}
\label{fig:reasoning_agent_prompt}
\end{figure*}

\begin{figure*}[t]
\centering
\begin{tcolorbox}[
    colback=white,
    colframe=purple!80,
    boxrule=1.5pt,
    arc=4pt,
    left=8pt,
    right=8pt,
    top=6pt,
    bottom=6pt,
    width=0.96\textwidth,
    title={\textbf{Verifier Agent Prompt}},
    fonttitle=\small\bfseries,
    coltitle=white,
    colbacktitle=purple!80
]
\small

\begin{tcolorbox}[
    colback=gray!10,
    colframe=gray!60,
    boxrule=1pt,
    arc=3pt,
    left=6pt,
    right=6pt,
    top=4pt,
    bottom=4pt
]
\textbf{System Prompt:}\\[0.3em]
You are a Verifier agent responsible for analyzing incorrect function calls and providing corrective feedback.

\vspace{0.3em}
\textbf{Query:} \texttt{\{hard\_query\}}

\vspace{0.3em}
\textbf{Model's Attempt:} \texttt{\{incorrect\_function\_call\}}

\vspace{0.3em}
\textbf{Ground Truth:} \texttt{\{correct\_function\_call\}}

\vspace{0.3em}
\textbf{Execution Result:} \texttt{\{execution\_result\}}
\end{tcolorbox}

\vspace{0.3em}

\begin{tcolorbox}[
    colback=orange!8,
    colframe=orange!60,
    boxrule=1pt,
    arc=3pt,
    left=6pt,
    right=6pt,
    top=4pt,
    bottom=4pt
]
\textbf{Your task:} Compare the model's attempt with the ground truth and identify:
\begin{enumerate}[leftmargin=1.5em, itemsep=0.2em]
    \item \textbf{Error Type:} Wrong tool selection, missing tool calls, incorrect parameters, wrong order, type mismatch, or missing dependencies
    \item \textbf{Root Cause:} Why did the model make this error? What logical step was missed? What dependency was not recognized?
    \item \textbf{Corrective Hint:} Specific guidance to fix this error without giving the answer directly
\end{enumerate}

\vspace{0.3em}

\textbf{Output format:}

\footnotesize\begin{verbatim}
{
  "error_type": "specific error category",
  "error_location": "which tool call or parameter",
  "root_cause": "explanation of why error occurred",
  "corrective_hint": "targeted guidance without 
                      revealing the answer",
  "should_reconsider": ["aspect1", "aspect2"]
}
\end{verbatim}

\vspace{0.2em}
\textbf{Example:}\\
\textit{Error:} Model called \texttt{buy\_tickets} directly without getting zipcodes first\\
\textit{Hint:} ``You attempted to purchase tickets directly, but the \texttt{buy\_tickets} tool requires zipcode parameters, not city names. Consider what information you need to obtain first before making the purchase.''
\end{tcolorbox}

\vspace{0.3em}

\begin{tcolorbox}[
    colback=red!8,
    colframe=red!60,
    boxrule=1pt,
    arc=3pt,
    left=6pt,
    right=6pt,
    top=4pt,
    bottom=4pt
]
\textbf{Remember:}
\begin{itemize}[leftmargin=1.5em, itemsep=0.2em]
    \item Do NOT give the answer directly---guide the model to discover the correct approach
    \item Focus on the logical reasoning gap, not just the technical error
    \item Hints should be specific enough to be helpful but general enough to require thinking
\end{itemize}
\end{tcolorbox}

\end{tcolorbox}
\caption{\textbf{Prompt template for the Verifier Agent}. Instructions for analyzing incorrect function calls and providing targeted corrective feedback.}
\label{fig:verifier_agent_prompt}
\end{figure*}

\begin{figure*}[t]
\centering
\begin{tcolorbox}[
    colback=white,
    colframe=purple!80,
    boxrule=1.5pt,
    arc=4pt,
    left=8pt,
    right=8pt,
    top=6pt,
    bottom=6pt,
    width=0.96\textwidth,
    title={\textbf{Reasoner Agent Prompt (Refinement Iteration)}},
    fonttitle=\small\bfseries,
    coltitle=white,
    colbacktitle=purple!80
]
\small

\begin{tcolorbox}[
    colback=gray!10,
    colframe=gray!60,
    boxrule=1pt,
    arc=3pt,
    left=6pt,
    right=6pt,
    top=4pt,
    bottom=4pt
]
\textbf{System Prompt:}\\[0.3em]
You are refining your previous attempt based on feedback.

\vspace{0.3em}
\textbf{Original Query:} \texttt{\{hard\_query\}}

\vspace{0.3em}
\textbf{Available Tools:} \texttt{\{tool\_descriptions\}}

\vspace{0.3em}
\textbf{Hint:} \texttt{\{advanced\_tool\_description\}}

\vspace{0.3em}
\textbf{Your Previous Attempt:}

\vspace{0.2em}
\footnotesize\begin{verbatim}
<think>
{previous_reasoning}
</think>
<tool_call>
{previous_function_call}
</tool_call>
\end{verbatim}

\vspace{0.2em}
\textbf{Feedback from Verifier:} \texttt{\{error\_diagnosis\_and\_hint\}}
\end{tcolorbox}

\vspace{0.3em}

\begin{tcolorbox}[
    colback=orange!8,
    colframe=orange!60,
    boxrule=1pt,
    arc=3pt,
    left=6pt,
    right=6pt,
    top=4pt,
    bottom=4pt
]
\textbf{Your task:}
\begin{enumerate}[leftmargin=1.5em, itemsep=0.2em]
    \item Carefully read the feedback
    \item Identify what went wrong in your previous attempt
    \item Revise your reasoning process
    \item Generate a corrected function call sequence
\end{enumerate}

\vspace{0.3em}

\textbf{Output format:}

\footnotesize\begin{verbatim}
<think>
[Updated reasoning addressing the feedback]
- What was wrong in my previous attempt?
- What does the feedback tell me?
- How should I adjust my approach?
- New execution plan
</think>

<tool_call>
[corrected_function_call_1, corrected_function_call_2, ...]
</tool_call>
\end{verbatim}

\end{tcolorbox}

\vspace{0.3em}

\begin{tcolorbox}[
    colback=red!8,
    colframe=red!60,
    boxrule=1pt,
    arc=3pt,
    left=6pt,
    right=6pt,
    top=4pt,
    bottom=4pt
]
\textbf{Remember:}
\begin{itemize}[leftmargin=1.5em, itemsep=0.2em]
    \item The feedback is designed to guide you toward the correct solution
    \item Focus on understanding WHY your previous attempt failed
    \item Each refinement should address the specific issues mentioned in feedback
    \item Maintain the same output format: \texttt{<think>} followed by \texttt{<tool\_call>}
\end{itemize}
\end{tcolorbox}

\end{tcolorbox}
\caption{\textbf{Prompt template for the Reasoner Agent (Refinement Iteration)}. Instructions for revising reasoning and function calls based on feedback from the Verifier.}
\label{fig:refinement_agent_prompt}
\end{figure*}

\label{sec:appendix}
\end{document}